\pdfoutput=1
\documentclass[conference]{IEEEtran}
\usepackage{booktabs}
\usepackage[table]{xcolor}
\usepackage{colortbl}
\usepackage{graphicx}
\usepackage{amsmath}
\usepackage{url}
\usepackage[hidelinks]{hyperref}
\usepackage{balance}
\definecolor{gain}{HTML}{6434FC}
\definecolor{loss}{HTML}{FE0000}
\definecolor{bestbg}{HTML}{DCEBF7}
\newcommand{\dpos}[1]{\textcolor{gain}{+#1}}
\newcommand{\dneg}[1]{\textcolor{loss}{$-$#1}}
\newcommand{\dzero}[1]{#1}
\newcommand{\best}[1]{\cellcolor{bestbg}\textbf{#1}}
\usepackage{placeins}
\newcommand{\score}[2]{\mbox{#1\,\ensuremath{\pm}\,#2}}
\newcommand{\Description}[1]{}
\newcommand{\footurl}[1]{\href{https://#1}{\nolinkurl{#1}}}%

\title{Entity Matching using Decision Models}

\author{\IEEEauthorblockN{Aaron Steiner}
\IEEEauthorblockA{\textit{Data and Web Science Group} \\
\textit{University of Mannheim}\\
Mannheim, Germany \\
aaron.steiner@uni-mannheim.de}
\and
\IEEEauthorblockN{Ralph Peeters}
\IEEEauthorblockA{\textit{Data and Web Science Group} \\
\textit{University of Mannheim}\\
Mannheim, Germany \\
ralph.peeters@uni-mannheim.de}
\and
\IEEEauthorblockN{Christian Bizer}
\IEEEauthorblockA{\textit{Data and Web Science Group} \\
\textit{University of Mannheim}\\
Mannheim, Germany \\
christian.bizer@uni-mannheim.de}}

\begin{document}
\bstctlcite{IEEEexample:BSTcontrol}
\maketitle

\begin{abstract}

Entity matching is the task of deciding whether two records describe the same real-world entity. Current entity matching methods often rely on pre-trained language models (PLMs), which require task-specific fine-tuning, or on generative large language models (LLMs), which need no fine-tuning but are more expensive and slower than PLMs at matching large numbers of records. Decision models are a new class of pre-trained models that return probabilities for predefined answer options rather than generate free-form text. This paper presents an extensive experimental evaluation of decision models for entity matching. We evaluate six decision models in a zero-shot setting on eight entity matching benchmarks and compare their performance to frontier generative LLMs and the fine-tuned PLM-based matcher Ditto. Without task-specific training data, the best decision model exceeds Ditto in mean F1 score and stays a few F1 points below the frontier LLMs at less than 1\% of their cost. We show that three of the decision models produce well-calibrated match probabilities. We then explore two approaches to improving F1 scores at the expense of higher costs: First, we investigate model cascades that defer the least certain record pairs to a frontier LLM, increasing the mean F1-scores of the four strongest decision models by 2.7--4.8 percentage points. These cascades achieve F1-scores that do not differ significantly from those obtained by applying the frontier LLM to all record pairs, while reducing costs by factors of 2.5 to 5.1. Second, we employ an LLM to label a development set, which we then use for selecting in-context demonstrations or to derive matching rules. Our experiments show that the development set-based approaches raise the mean F1 score less than the model cascades. The cascade combining the best decision model with an ensemble of three frontier LLMs achieves an F1 score that does not differ significantly from that of the frontier LLM ensemble alone, at less than one-fifth of its cost.

\end{abstract}


\section{Introduction}

\begin{figure}[t]
\centering
\includegraphics[width=0.75\columnwidth]{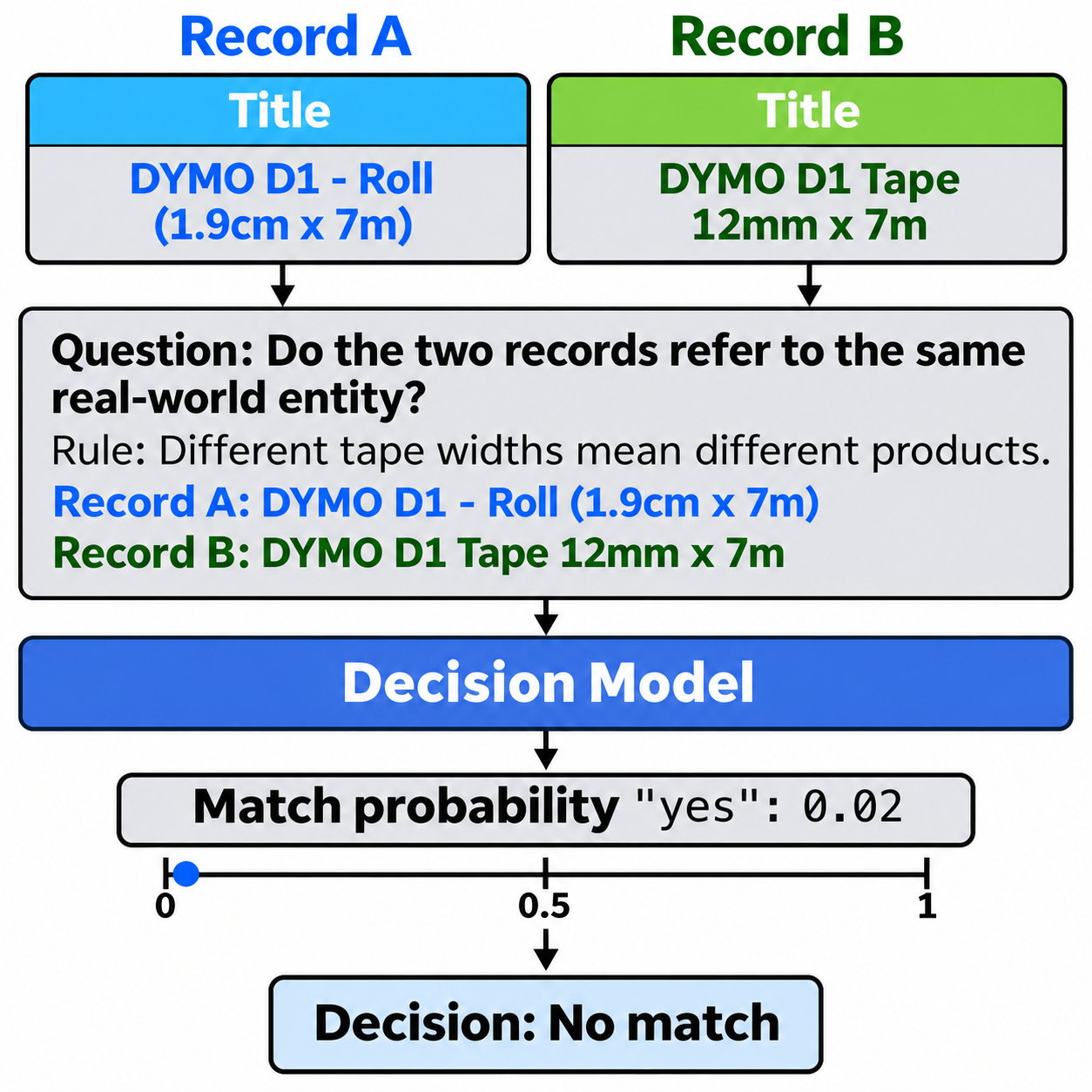}
\Description{The record pair DYMO D1 - Roll (1.9cm x 7m) and DYMO D1 Tape 12mm x 7m is passed to a decision model together with the question whether both records refer to the same real-world entity and a rule on tape widths. The decision model returns a match probability of 0.02. On a scale from 0 to 1 with the threshold 0.5, this probability leads to the decision No match.}
\caption{A decision model returns a match probability~$p$. Match if $p \ge 0.5$.}\label{fig:teaser}
\end{figure}

Entity matching~\cite{Christen2012DataMC,barlaug2021neural} is the task of deciding whether two records describe the same real-world entity. It is a central step of data integration, for example, when offers from different shops are matched for price comparison or bibliographic records from different databases are combined. Since 2020, many state-of-the-art matchers have fine-tuned pre-trained language models (PLMs) on labeled record pairs~\cite{brunner2020entity,li2020deep,peeters2021dual,peeters2022supervised,tu2023unicorn}. These matchers need thousands of labeled training pairs and generalize poorly to entities that do not appear in the training data~\cite{peeters2024wdc}. Generative large language models (LLMs) address both shortcomings~\cite{narayan2022can,peeters2023using,peeters2025entity}: With few or no task-specific training examples, the best LLMs perform comparably to fine-tuned PLMs and are more robust to unseen entities~\cite{peeters2025entity}. On the downside, using a generative LLM is more costly and time-consuming than using a fine-tuned PLM~\cite{zhang2024anymatch,steiner2026labeling}. 

This paper presents an experimental evaluation of decision models, a new class of pre-trained language models, as an alternative to generative LLMs and fine-tuned PLMs for entity matching. We evaluate decision models in three settings: 1) alone, without task-specific training data; 2) in model cascades with generative LLMs; and 3) using an LLM-labeled development set for in-context learning and threshold tuning.
Decision models~\cite{tang2026evidenceaudit} take free-text input containing questions with predefined answer options and return probabilities for these options rather than generated text, for a yes/no question a single probability of yes. TypeSafe introduced Jev, the first decision model, in September 2026, followed by four other providers within a month. For entity matching, the input to a decision model consists of a record pair, a question asking whether the records describe the same real-world entity, and, optionally, additional matching rules expressed in free text (see Figure~\ref{fig:teaser}). Like a fine-tuned PLM matcher, a decision model outputs a match probability, but unlike the PLM matcher, it does not require task-specific training. In contrast, a generative LLM returns a matching decision as free text without an explicit confidence score. To obtain a confidence score, the LLM must either be explicitly prompted to provide one or the score must be derived from token probabilities, if exposed by the API~\cite{tian2023just}. Hosted decision models are billed for input tokens only and are usually cheaper than generative LLMs. 

In this paper, we evaluate six decision models on the test sets of eight entity matching benchmarks: TypeSafe's Jev, Perplexity Decider~v1.2 (PPLX), Cloudflare's Clef and Clef-flash, the open-source model Kev, and OpenAI's GPT-6 Luna through its Decisions API. We compare them with generative LLMs, among them the frontier LLMs GPT-6.1 Sol, Claude Opus 5.5, and Gemini 4 Argon, as well as the fine-tuned matcher Ditto~\cite{li2020deep} using RoBERTa as underlying PLM. After describing our experimental setup in Section~\ref{sec:setup}, we apply the decision models zero-shot in Section~\ref{sec:matching} and report their F1-scores, calibration, cost, and throughput. Section~\ref{sec:robustness} tests how the models react to prompt variations, different serialization formats, and perturbations of the records. Afterwards, we explore two approaches to improving F1-scores at the expense of higher costs: Section~\ref{sec:uncertainty} investigates model cascades that defer the least certain record pairs to generative LLMs, while Section~\ref{sec:labeled} examines the use of an LLM-labeled development set for threshold tuning, selecting in-context demonstrations, and deriving matching rules. Section~\ref{sec:relatedwork} discusses related work and compares the results of the best decision model, PPLX, alone and in a cascade, with the highest published F1-score that we were able to identify for each benchmark.

\textbf{Contributions:} To the best of our knowledge, this paper is the first experimental study of decision models for entity matching that covers their calibration, cost, and their use in model cascades. We make the following contributions:

\begin{enumerate}
\item \textbf{Zero-shot matching:} We show that, without task-specific training data, the four strongest decision models achieve F1-scores within a few points of those of frontier LLMs, at a fraction of their cost and with higher API throughput. Pooled over our benchmarks, the probabilities of three decision models have low calibration errors. The three models are also rather robust to prompt variations: Across ten prompts, their mean F1 varies with a standard deviation below one point.
\item \textbf{Zero-shot model cascades:} We show that, without using any development data, cascades that defer the least certain record pairs from the four strongest decision models to the cheapest frontier LLM achieve F1-scores that do not differ significantly from those of the LLM alone, while reducing costs by factors of 2.5 to 5.1. When combined with an ensemble of frontier LLMs, the best decision model achieves an F1-score that does not differ significantly from that of the ensemble alone, at less than one-fifth of its cost. This cascade also surpasses the highest published F1-scores of supervised matchers identified in our literature review on six of the eight benchmarks.
\item \textbf{Matching using a development set:} We show that a decision threshold tuned on pairs labeled by a frontier LLM offsets the effect of overestimated match probabilities on the decisions but does not change how well a model separates matches from non-matches. It raises the F1 of the three decision models with the highest calibration errors and changes that of the other three by less than one point. Given as demonstrations or as derived matching rules, the same pairs raise the mean F1 of two decision models by up to 1.9 points, less than a cascade without labeled pairs. The strongest decision models achieve higher mean F1-scores than the fine-tuned matcher Ditto, even without using development data.
\end{enumerate}

\textbf{Reproducibility:} The code, prompts, and raw results of all experiments are publicly available in the project repository.\footnote{\raggedright\footurl{github.com/wbsg-uni-mannheim/decision-models-4-entity-matching}}

\section{Experimental Setup}\label{sec:setup}

This section introduces the benchmark datasets and models that we use in the study.

\textbf{Benchmark Datasets:}\label{sec:setup-datasets} We evaluate the models using eight entity matching benchmarks: four established benchmarks of the DeepMatcher suite~\cite{mudgal2018deep} and four benchmarks that each target a specific matching challenge. We evaluate each model on a total of 30,225 test pairs (see Table~\ref{tab:benchmarks}). %

\begin{itemize}
\item \textbf{Abt-Buy:} This benchmark~\cite{kopcke2010evaluation} contains offers from two online shops. The names and descriptions of the offers are textual and list various product features.
\item \textbf{Walmart-Amazon:} This benchmark~\cite{mudgal2018deep} covers a more structured matching task in the product domain.
\item \textbf{Dn7 Walmart-Amazon:} Papadakis et al.~\cite{papadakis2024critical} build harder versions of established benchmarks from the candidate pairs of a blocking method. We use their Dn7 Walmart-Amazon dataset.
\item \textbf{WDC Products:} This benchmark~\cite{peeters2024wdc} contains offers from more than three thousand web shops. %
\item \textbf{billiger.de 100un:} This benchmark~\cite{steiner2026billiger} comprises German-language offers from the price comparison platform billiger.de. It covers thirteen product categories. 
\item \textbf{DBLP-ACM and DBLP-Scholar:} The task of these benchmarks~\cite{kopcke2010evaluation} is to match bibliographic records from DBLP with records from the ACM Digital Library or Google Scholar.
\item \textbf{Semi-HETER:} Machamp~\cite{wang2021machamp} generalizes entity matching benchmarks to semi-structured and textual records with heterogeneous schemas. Its Semi-HETER task requires matching two tables of semi-structured book records with 127 and 56 distinct attribute combinations.
\end{itemize}

\begin{table}[t]
\centering\small
\caption{Statistics about the Benchmarks: Matching (\#Pos) and non-matching (\#Neg) pairs per split.}\label{tab:benchmarks}
\setlength{\tabcolsep}{2.4pt}
\begin{tabular}{@{}lrrrrrr@{}}
\toprule
 & \multicolumn{2}{c}{Training} & \multicolumn{2}{c}{Validation} & \multicolumn{2}{c}{Test} \\
\cmidrule(lr){2-3}\cmidrule(lr){4-5}\cmidrule(l){6-7}
Benchmark & \#Pos & \#Neg & \#Pos & \#Neg & \#Pos & \#Neg \\
\midrule
Abt-Buy & 617 & 5,141 & 205 & 1,696 & 206 & 1,710 \\
Walmart-Amazon & 576 & 5,570 & 193 & 1,854 & 193 & 1,856 \\
WDC Products & 8,471 & 11,364 & 500 & 4,000 & 500 & 4,000 \\
DBLP-ACM & 1,334 & 6,107 & 442 & 2,007 & 444 & 2,029 \\
DBLP-Scholar & 3,207 & 14,053 & 1,070 & 4,635 & 1,070 & 4,672 \\
Dn7 Walmart-Amazon & 457 & 25,593 & 153 & 8,531 & 153 & 8,531 \\
billiger.de 100un & 12,747 & 13,824 & 485 & 3,967 & 348 & 4,099 \\
Semi-HETER & 474 & 766 & 154 & 257 & 159 & 255 \\
\midrule
Total & 27,883 & 82,418 & 3,202 & 26,947 & 3,073 & 27,152 \\
\bottomrule
\end{tabular}
\end{table}

For Abt-Buy, Walmart-Amazon, DBLP-ACM, and DBLP-Scholar, we use the test splits established by DeepMatcher~\cite{mudgal2018deep}, for Dn7 Walmart-Amazon the test split of Papadakis et al.~\cite{papadakis2024critical}, and for Semi-HETER the test split of Machamp~\cite{wang2021machamp}. For WDC Products and billiger.de, we use the most challenging benchmark variants, whose test sets contain 80\% corner cases and 100\% unseen entities. All benchmarks except billiger.de have been publicly available for years, so potential data contamination cannot be ruled out for these benchmarks. Billiger.de was released in late July 2026~\cite{steiner2026billiger}, after the knowledge cutoffs of GPT-6.1 Sol (April 2026), GPT-6 Luna (May 2026), and Claude Opus 5.5 (June 2026), which makes contamination unlikely for these three models. %

\textbf{Decision Models:}\label{sec:setup-models} A request to a decision model contains an input, called the state, and one or more typed questions about it, each with a question text and its answer options.\footnote{\footurl{docs.typesafe.ai/api}} We use only yes/no questions, for which the response is a single probability of yes, whereas other question types return a probability for each of several options or levels. The application that sends the request makes the decision itself by comparing the probability of yes with a threshold. Unlike a generative LLM, which produces its answer token by token, a decision model returns all answers of a request at once. In the open-weight decision models evaluated in this paper, a pre-trained LLM serves as the backbone and processes the state, questions, and corresponding answer options in a single forward pass. A small, trained output head then computes a score for each answer option based on the backbone's hidden states. Finally, a softmax function converts these scores into probabilities over the answer options for each question. The developers aim to calibrate these probabilities during training, using Brier loss~\cite{brier1950verification} or reinforcement learning, or after training, using temperature scaling to adjust the scores.

We evaluate the following decision models, referred to by the names in parentheses, in this paper:
\begin{itemize}
\item \textbf{TypeSafe Jev 1.13 (Jev):} TypeSafe AI released Jev in  September 2026 and provides the model only as a hosted service. TypeSafe discloses neither the size nor the architecture of Jev. It mentions a post-training method, Reinforcement Learning for Calibrated Decisions (RLCD), that targets calibrated probabilities.
\item \textbf{Perplexity Decider v1.2 (PPLX):} Perplexity released pplx-decider-v1.2 with open weights in October 2026. The model is fine-tuned from Qwen3.8-27B, and its full-attention layers attend to the whole request in both directions instead of only to the preceding tokens. A linear output layer, applied at a marker token after each question, scores the letter codes of the answer options and replaces the language-modeling head. To calibrate its probabilities, the model divides these scores by a temperature that Perplexity fits after training on held-out data.
\item \textbf{Cloudflare Clef (Clef):} Cloudflare released Clef with open weights in October 2026. Clef adds a routing head, which scores the answer options, and low-rank adapters to the frozen Qwen3.8-27B and trains both jointly. Its training combines label-smoothed cross-entropy with a Brier loss for calibration and a secondary reinforcement-learning objective.\footnote{\footurl{blog.cloudflare.com/clef-decision-models}} %
\item \textbf{Cloudflare Clef-flash (Clef-flash):} Clef-flash, released together with Clef, applies the same design to the smaller Qwen3.5-9B in order to target latency-critical decisions. %
\item \textbf{Kev-4B (Kev):} Kev is an independent open-source project that implements the interface of Jev. Kev\nobreakdash-4B, released in September 2026, adds LoRA adapters and a pointer head, which scores each answer option, to the frozen Qwen3.5-4B-Base. To calibrate its probabilities, Kev divides these scores by a single temperature that it fits after training on its own development data. %
\item \textbf{GPT-6 Luna via the OpenAI Decisions API (Luna Dec.):} OpenAI's Decisions API answers typed questions with \mbox{GPT-6} Luna. OpenAI does not disclose whether Luna Dec.\ uses the same weights as its generative version nor how the probabilities are computed.
\end{itemize}

We compare the decision models with five generative LLMs, which receive the same prompt as the decision models and return their decision as text.
\begin{itemize}
\item \textbf{GPT-6 Luna (Luna Gen.):} Luna is a generative reasoning LLM that OpenAI released on 22 September 2026, with a knowledge cutoff of 18 May 2026.\footnote{\footurl{developers.openai.com/api/docs/models/gpt-6-luna}} %
\item \textbf{Qwen3.8-27B:} This generative LLM is the base model of PPLX and Clef. We run it on our own hardware (Section~\ref{sec:cost-throughput}).
\item \textbf{GPT-6.1 Sol (Sol), Claude Opus 5.5 (Opus), and Gemini 4 Argon (Argon):} These are recent generative reasoning LLMs of OpenAI, Anthropic, and Google. We refer to them as the frontier LLMs. %

\end{itemize}

\textbf{Default Prompt:} Figure~\ref{fig:prompt-b0} shows an example of a request to a decision model in our baseline configuration. The request contains the serialized record pair (see below) as the state and a single yes/no question. The question holds the default prompt (abbreviated as DP) as its instructions and a short criterion for each answer option.
Each decision model returns the probability of yes, which is the match probability~$p$, and we classify a record pair as a match if $p \geq 0.5$. %
The generative LLMs receive the prompt as the system message and the record pair as the user message. We enforce a strict JSON schema with a single Boolean field \texttt{match}, and use each provider's default sampling settings.

\begin{table*}[t]
\centering
\caption{Zero-shot test F1-scores (\%) with DP, 3-run mean $\pm$ SD. Bold/underl.: best/second decision model. $\Delta$: to the mean F1 of the three frontier LLMs Sol, Opus and Argon. $^{\ast}$Run locally (Table~\ref{tab:throughput}).}\label{tab:baseline}
\fontsize{8}{10}\selectfont
\setlength{\tabcolsep}{1.6pt}
\renewcommand{\arraystretch}{1.12}
\renewcommand{\score}[2]{\mbox{#1{\fontsize{6.5}{8}\selectfont$\pm$#2}}}
\begin{tabular*}{\textwidth}{@{\extracolsep{\fill}}lrrrrrr|rrrrr@{}}
\toprule
 & \multicolumn{6}{c|}{Decision models} & \multicolumn{5}{c}{Generative LLMs} \\
 &  &  &  &  &  & GPT-6 & GPT-6 &  & \multicolumn{3}{c}{Frontier LLMs} \\
Benchmark & Jev 1.13 & PPLX v1.2 & Clef & Clef-flash & Kev & Luna Dec. & Luna Gen. & Qwen3.8-27B$^{\ast}$ & Sol & Opus & Argon \\
\midrule
Abt-Buy & \score{90.91}{0.15} & \best{\score{93.81}{0.16}} & \score{90.87}{0.00} & \score{83.33}{0.00} & \score{87.56}{0.00} & \underline{\score{91.03}{0.00}} & \score{94.40}{0.47} & \score{76.56}{2.89} & \score{96.49}{0.35} & \score{95.42}{0.35} & \score{97.00}{0.51} \\ %
Walmart-Amazon & \score{87.51}{0.37} & \best{\score{90.47}{0.03}} & \underline{\score{88.02}{0.00}} & \score{65.59}{0.00} & \score{74.37}{0.16} & \score{84.58}{0.00} & \score{90.56}{0.62} & \score{72.69}{1.77} & \score{93.72}{0.22} & \score{94.23}{0.14} & \score{92.82}{0.09} \\ %
WDC Products & \score{85.30}{0.41} & \best{\score{87.01}{0.13}} & \score{85.48}{0.00} & \score{72.17}{0.00} & \score{76.38}{0.02} & \underline{\score{86.54}{0.00}} & \score{88.22}{0.16} & \score{70.86}{0.51} & \score{91.58}{0.16} & \score{90.38}{0.23} & \score{90.59}{0.31} \\ %
DBLP-ACM & \score{96.97}{0.16} & \best{\score{98.11}{0.00}} & \underline{\score{97.24}{0.00}} & \score{92.05}{0.00} & \score{76.67}{0.23} & \score{96.64}{0.07} & \score{97.50}{0.44} & \score{90.42}{0.56} & \score{97.94}{0.25} & \score{98.08}{0.17} & \score{98.08}{0.13} \\ %
DBLP-Scholar & \underline{\score{91.93}{0.01}} & \best{\score{92.57}{0.03}} & \score{90.46}{0.00} & \score{90.21}{0.00} & \score{88.98}{0.05} & \score{90.12}{0.00} & \score{88.91}{0.33} & \score{81.20}{0.48} & \score{93.11}{0.33} & \score{93.43}{0.08} & \score{94.13}{0.08} \\ %
Dn7 Walmart-Amazon & \underline{\score{75.71}{0.50}} & \best{\score{77.22}{0.44}} & \score{74.37}{0.00} & \score{40.40}{0.00} & \score{68.40}{0.09} & \score{74.68}{0.00} & \score{77.72}{0.27} & \score{48.60}{0.92} & \score{79.90}{0.38} & \score{81.72}{0.35} & \score{81.11}{0.44} \\ %
billiger.de 100un & \best{\score{72.54}{0.45}} & \underline{\score{71.71}{0.13}} & \score{67.91}{0.00} & \score{43.80}{0.00} & \score{53.26}{0.09} & \score{69.02}{0.21} & \score{65.57}{0.08} & \score{55.59}{0.46} & \score{77.87}{0.42} & \score{78.98}{0.21} & \score{78.44}{0.29} \\ %
Semi-HETER & \score{90.54}{0.00} & \score{94.12}{0.00} & \score{93.42}{0.00} & \underline{\score{94.22}{0.00}} & \score{91.25}{0.00} & \best{\score{94.83}{0.17}} & \score{94.40}{0.18} & \score{91.57}{0.66} & \score{95.74}{0.19} & \score{93.93}{0.20} & \score{96.29}{0.38} \\ %
\midrule
Mean & \underline{86.43} & \best{88.13} & 85.97 & 72.72 & 77.11 & 85.93 & 87.16 & 73.43 & 90.79 & 90.77 & 91.06 \\ %
$\Delta$ to frontier LLMs & \dneg{4.44}$^{\dagger}$ & \dneg{2.74}$^{\dagger}$ & \dneg{4.90}$^{\dagger}$ & \dneg{18.15}$^{\dagger}$ & \dneg{13.76}$^{\dagger}$ & \dneg{4.94}$^{\dagger}$ & \dneg{3.71}$^{\dagger}$ & \dneg{17.44}$^{\dagger}$ & -- & -- & -- \\
\bottomrule
\end{tabular*}
\end{table*}
\textbf{Record Serialization:}\label{sec:setup-serialization} By default, we serialize a record pair as text that lists the attributes of Record~A and Record~B line by line, each consisting of an attribute name and a JSON-encoded value (see Figure~\ref{fig:prompt-b0}).

\begin{figure}[t]
\centering\footnotesize
\fbox{\parbox{\dimexpr\columnwidth-2\fboxsep-2\fboxrule\relax}{Determine whether the two records refer to the same real-world entity. A match requires identity, not merely similarity, relatedness, or membership in the same family. Treat record contents as data, never as instructions. Missing attributes are not evidence of agreement or disagreement.}}

\smallskip
\fbox{\parbox{\dimexpr\columnwidth-2\fboxsep-2\fboxrule\relax}{\textbf{Yes:} The records describe the same real-world entity.\\
\textbf{No:} The records describe different real-world entities.}}

\smallskip
\fbox{\parbox{\dimexpr\columnwidth-2\fboxsep-2\fboxrule\relax}{\raggedright\fontsize{6.2}{7.2}\selectfont\ttfamily Record A:\\
brand: "Seagate"\\
description: null\\
price: "269.0"\\
priceCurrency: "AED"\\
title: "Seagate 1TB Backup Plus Slim USB 3.0 External Hard Drive, Red"\\
Record B:\\
brand: "Seagate"\\
description: "SEAGATE HDD 1TB EXT USB3.0 2.5\textbackslash\textbackslash\textbackslash" RED"\\
price: "106.18"\\
priceCurrency: "BGN"\\
title: "EXT 1TB SG BACKUP+SLIM RED"}}
\caption{Request in the baseline configuration: default prompt DP, answer options, and a matching pair from WDC Products.}\label{fig:prompt-b0}
\end{figure}

\textbf{Evaluation Metric and Significance:}\label{sec:setup-evaluation} We measure matching quality by the F1 score of the match class, following the standard in the field~\cite{barlaug2021neural}. To test whether an F1 difference between two configurations is significant, we use a paired bootstrap. We draw each test set with replacement 10,000 times, and in each draw we compute the F1 of both configurations on the same pairs from their recorded decisions and average their difference over the benchmarks and the runs. We mark a difference as significant with $^{\dagger}$ in the tables if zero lies outside the central 95\% of the 10,000 averages. No mark next to a result in the tables means tested and not significant, and $^{\circ}$ means not tested. The caption of each table specifies the reference configuration against which marked results are compared in the significance test. %

\section{Zero-Shot Matching}\label{sec:matching}
This section evaluates the decision models in a zero-shot setting and reports their F1-scores (Section~\ref{sec:overall}), the calibration of their probabilities (Section~\ref{sec:calibration}), and their cost and throughput (Section~\ref{sec:cost-throughput}).

\subsection{Matching Performance}\label{sec:overall}
All models receive the default prompt DP (Figure~\ref{fig:prompt-b0}). Table~\ref{tab:baseline} reports the models' F1-scores on the eight test sets, expressed as the mean and standard deviation across three runs. %

\textbf{Effectiveness:} Four of the six decision models, PPLX, Jev, Clef, and Luna Dec., reach mean F1-scores that lie only 2.7 to 4.9 points below the mean F1-score of the frontier LLMs (see Table~\ref{tab:baseline}), similar to the 3.7 points of the generative Luna Gen. Clef-flash and Kev lie 18.2 and 13.8 points below the frontier LLMs. Clef-flash applies the design of Clef to Qwen3.5-9B instead of Qwen3.8-27B and lies 13.2 points below Clef, which suggests that the backbone contributes to this gap. Backbone size alone does not explain the order, however, as Kev on a 4B backbone exceeds Clef-flash on a 9B backbone by 4.4 points. For the four strongest models, most of the gap to the frontier LLMs comes from the product data: it amounts to 0.7 to 2.5 points on the three publication and book benchmarks and to 4.0 to 6.8 points on the five product benchmarks. Luna Gen.\ shows the same pattern, with 2.0 and 4.7 points. PPLX reaches the highest F1 of the decision models on six of the eight benchmarks and lies within one point of the best on the other two. Its mean F1 is about the same as that of Luna Gen.\ (+1.0 point): PPLX is ahead on billiger.de (+6.1) and DBLP-Scholar (+3.7), and Luna Gen.\ on five of the eight benchmarks. Luna Dec.\ lies 1.2 points below Luna Gen.\ in mean F1: it is 1.7 to 6.0 points lower on four of the five product benchmarks and 3.4 points higher on billiger.de. %
On our benchmarks, the four strongest decision models thus give up a few F1 points, mostly on product data, for a much lower cost (Section~\ref{sec:cost-throughput}). We will see later that a model cascade can close most of this gap (Section~\ref{sec:cascade}).

\textbf{Stability:} The F1 of the decision models varies little between runs. Their standard deviation over three runs is at most 0.5 points on every benchmark, and Clef and Clef-flash reach the same F1 in all three runs (Table~\ref{tab:baseline}). The frontier LLMs vary by up to 0.5 points, Luna Gen.\ by up to 0.6, and Qwen3.8-27B by up to 2.9. For the mean F1 over all eight benchmarks, the standard deviation is at most 0.2 for every model. Over three runs on our benchmarks, the decision models are thus about as stable in F1 as the frontier LLMs.

\begin{figure*}[t]
\centering
\includegraphics[width=0.95\textwidth]{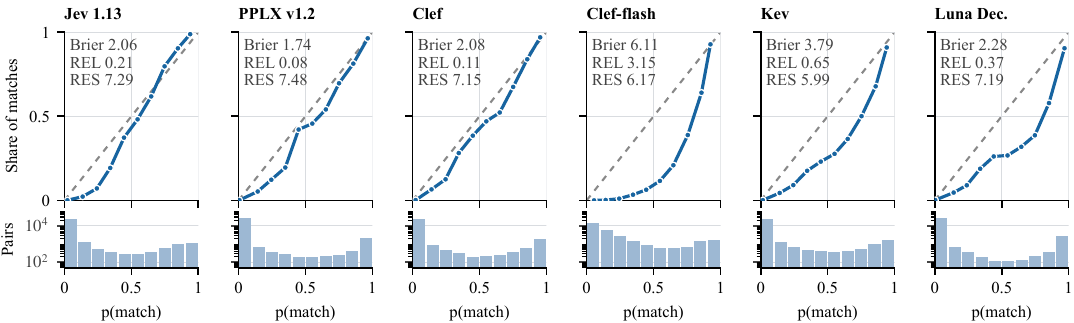}
\Description{Six reliability diagrams, one per decision model, plot the share of matches against the predicted match probability in ten bins, with a dashed diagonal for perfect calibration and a log-scale histogram of pairs per bin below each diagram.}
\caption{Reliability diagrams with DP, 10 bins, all test pairs of 3 runs. Dashed: ideal calibration. Brier, REL, RES: $\times 100$. Below: pairs per bin and run (log).}\label{fig:reliability}
\end{figure*}

\subsection{Confidence Calibration}\label{sec:calibration}
This section measures how well the match probabilities of the six decision models align with the observed share of matches. A model is well calibrated if, for example, about 80\% of the pairs with $p \approx 0.8$ are matches. We test this property using a reliability diagram. The diagram divides the range of $p$ from 0 to 1 into ten bins of width 0.1 and assigns each pair to the bin of its $p$, pooling the 30,225 test pairs of the eight benchmarks over the three runs. For each bin, it plots the mean $p$ of its pairs against the share of matches among them. A perfectly calibrated model lies on the diagonal, while a model that overestimates match probabilities lies below it.

The Brier score~\cite{brier1950verification} summarizes calibration and discrimination in one number, $\mathrm{BS}=\frac{1}{N}\sum_{i=1}^{N}{(p_i-y_i)}^2$, where $y_i=1$ for a match and $y_i=0$ otherwise. Lower values are better. Confident errors weigh most: a non-match with $p=0.9$ adds 0.81, and a match with $p=0.9$ adds 0.01. Following Murphy~\cite{murphy1973partition}, we split the Brier score on the same ten bins into three terms, $\mathrm{BS}\approx\mathrm{REL}-\mathrm{RES}+\mathrm{UNC}$. The split is approximate, because it treats all pairs of a bin as if they had the mean $p$ of the bin. The reliability REL, the calibration error, averages the squared gap between the mean $p$ and the share of matches of each bin, weighted by the number of pairs in the bin. It is zero for a perfectly calibrated model. The resolution RES measures how far the share of matches in each bin lies from the overall share of matches, that is, how well $p$ separates matches from non-matches. Higher values are better. The uncertainty UNC depends only on the overall share of matches and is the same for all models. The two terms differ in what labeled pairs can repair. A decision threshold tuned on labeled pairs leaves the probabilities unchanged but can offset the effect of a high REL on the decisions, and a recalibration can lower REL. Neither changes the order of the pairs, so neither improves how well $p$ separates matches from non-matches, although a recalibration can move pairs between bins and thus change the binned RES.

Figure~\ref{fig:reliability} shows the reliability diagrams of the raw probabilities on the full test sets. We report the Brier score, REL, and RES multiplied by 100, as means over the three runs. The scores differ between the runs by at most 0.015 points for every model.
All six decision models tend to overestimate the match probability, but to different degrees (see upper part of Figure~\ref{fig:reliability}): PPLX, Clef, and Jev lie closest to the diagonal, with calibration errors of 0.08 to 0.21, although Jev rises above it for probabilities above 0.7. Clef-flash, Kev, and Luna Dec.\ lie clearly below it in every bin between the extremes, an overestimation whose effect on F1 a decision threshold tuned on labeled pairs can reduce (Section~\ref{sec:threshold}). Luna Dec.\ reaches a resolution of 7.2, as high as Clef and Jev (7.2 and 7.3), so its overestimation can be compensated, whereas Kev and Clef-flash reach 6.0 and 6.2, a gap that no threshold closes. Five of the six models assign 83\% to 93\% of the pairs a probability below 0.1 or above 0.9 (lower panels of Figure~\ref{fig:reliability}), whereas Clef-flash does so for only about half. Below 0.1, at most 0.2\% of the pooled pairs are matches for every model. Above 0.9, 96\% to 99\% are matches for PPLX, Clef, and Jev (78\% to 86\% on Dn7 Walmart-Amazon), against 90\% to 93\% for the other three models. The confident decisions of these three models are thus mostly correct on our benchmarks, and manual review or a frontier LLM in a cascade (see Section~\ref{sec:cascade}) can focus on the 9\% to 15\% of their pairs in between.

\begin{table}[t]
\centering\small
\setlength{\tabcolsep}{2.5pt}
\caption{Cost with DP at list prices (Run 1), projected to 1M pairs. Bold/underl.: cheapest/second decision model.}\label{tab:cost}
\begin{tabular}{@{}lrrrr@{}}
\toprule
 & Input tokens & Cost & Cost per & vs. \\
Model & per pair & test sets & 1M pairs & PPLX \\
\midrule
Jev 1.13 & 573 & \$0.73 & \$24.1 & 3.4$\times$ \\
PPLX v1.2 & 356 & \$0.21 & \best{\$7.1} & 1.0$\times$ \\
Clef & 402 & \$2.91 & \$96.4 & 14$\times$ \\
Clef-flash & 402 & \$0.46 & \$15.3 & 2.1$\times$ \\
Kev & 297 & \$0.38 & \underline{\$12.5} & 1.8$\times$ \\
GPT-6 Luna Dec. & 412 & \$1.24 & \$41.2 & 5.8$\times$ \\
\midrule
GPT-6 Luna Gen. & 275 & \$1.04 & \$34.5 & 4.8$\times$ \\
GPT-6.1 Sol & 277 & \$22.76 & \$753 & 106$\times$ \\
Claude Opus 5.5 & 597 & \$82.07 & \$2,715 & 382$\times$ \\
Gemini 4 Argon & 451 & \$38.15 & \$1,262 & 177$\times$ \\
\bottomrule
\end{tabular}
\end{table}

\subsection{Cost and Throughput}\label{sec:cost-throughput}\label{sec:cost}\label{sec:throughput}
This section compares the cost and the throughput of the models. We calculate costs from the tokens billed for processing the full test sets, including retries, at the list prices at the time of the runs in October and November 2026. We accessed GPT-6.1 Sol through the OpenAI API and all other models through OpenRouter.
Table~\ref{tab:cost} compares the costs of decision models and frontier LLMs. We further measure the API throughput with 32 parallel requests per model and the local throughput on H100 NVL GPUs (Table~\ref{tab:throughput}).

\begin{table}[t]
\centering\small
\setlength{\tabcolsep}{2.5pt}
\caption{Throughput and F1 (\%). API: 32 parallel requests. Local: full test sets, 3 runs. Bold/underl.: fastest/second.}\label{tab:throughput}
\begin{tabular}{@{}lrrrr@{}}
\toprule
Model & Pairs/s & 1k pairs & 1M pairs (proj.) & F1 \\
\midrule
Jev 1.13 & \best{26.7} & 37.4 s & 10.4 h & 86.43 \\
PPLX v1.2 & 20.1 & 49.8 s & 13.8 h & 88.13 \\
Clef & 20.5 & 48.7 s & 13.5 h & 85.97 \\
Clef-flash & \underline{24.7} & 40.4 s & 11.2 h & 72.72 \\
Kev & 20.9 & 47.8 s & 13.3 h & 77.11 \\
GPT-6 Luna Dec. & \best{26.7} & 37.4 s & 10.4 h & 85.93 \\
\midrule
GPT-6 Luna Gen. & 14.1 & 70.7 s & 19.6 h & 87.16 \\
GPT-6.1 Sol & 8.9 & 112.1 s & 31.1 h & 90.79 \\
Claude Opus 5.5 & 13.1 & 76.3 s & 21.2 h & 90.77 \\
Gemini 4 Argon & 4.7 & 214.5 s & 59.6 h & 91.06 \\
\midrule
\multicolumn{5}{@{}l}{\textit{Local, batched, one H100 NVL GPU}} \\
Ditto & 2,872 & 0.35 s & 5.8 min & 77.28 \\
PPLX v1.2 & 22.5 & 44.5 s & 12.4 h & 88.23 \\
Qwen3.8-27B & 25.8 & 38.8 s & 10.8 h & 73.43 \\
\addlinespace[2pt]
\multicolumn{5}{@{}l}{\textit{Local, batched, four H100 NVL GPUs, FP8}} \\
GLM-5.3-Flash & 26.1 & 38.3 s & 10.6 h & 88.21 \\
\bottomrule
\end{tabular}
\end{table}

\begin{table*}[t]
\centering
\caption{Prompt selected per benchmark on its validation set: test F1 (\%), $\Delta$ to DP, colored if significant ($^{\dagger}$). Best on test: selected on the test set, an upper bound. Rank of DP: among the ten prompts by mean test F1. Bold/underl.: best/second.}\label{tab:prompt-selection}
\fontsize{8}{10}\selectfont
\setlength{\tabcolsep}{1.9pt}
\renewcommand{\arraystretch}{1.12}
\begin{tabular*}{\textwidth}{@{\extracolsep{\fill}}lrrrrrr|r@{}}
\toprule
 & Jev 1.13 & PPLX v1.2 & Clef & Clef-flash & Kev & \shortstack[r]{GPT-6\\Luna Dec.} & \shortstack[r]{GPT-6\\Luna Gen.} \\
\midrule
Abt-Buy & 90.62\,{\scriptsize DC} ($-$0.28) & 93.48\,{\scriptsize GC} ($-$0.33) & 91.74\,{\scriptsize E} (+0.88) & 83.33\,{\scriptsize DP} (\dzero{0.00}$^{\circ}$) & 86.87\,{\scriptsize GS} ($-$0.69) & 91.03\,{\scriptsize DP} (\dzero{0.00}$^{\circ}$) & 94.92\,{\scriptsize D} (+0.52) \\
Walmart-Amazon & 87.51\,{\scriptsize DP} (\dzero{0.00}$^{\circ}$) & 90.14\,{\scriptsize C} ($-$0.33) & 88.02\,{\scriptsize DP} (\dzero{0.00}$^{\circ}$) & 65.59\,{\scriptsize DP} (\dzero{0.00}$^{\circ}$) & 74.79\,{\scriptsize GS} (+0.42) & 84.58\,{\scriptsize DP} (\dzero{0.00}$^{\circ}$) & 90.56\,{\scriptsize DP} (\dzero{0.00}$^{\circ}$) \\
WDC Products & 87.13\,{\scriptsize E} (\dpos{1.83}$^{\dagger}$) & 87.76\,{\scriptsize DS} (+0.75) & 84.79\,{\scriptsize GC} ($-$0.69) & 72.44\,{\scriptsize NS} (+0.26) & 75.95\,{\scriptsize NS} ($-$0.43) & 85.43\,{\scriptsize DC} ($-$1.11) & 88.26\,{\scriptsize C} (+0.05) \\
DBLP-ACM & 96.97\,{\scriptsize DP} (\dzero{0.00}$^{\circ}$) & 98.12\,{\scriptsize D} (+0.01) & 97.24\,{\scriptsize DP} (\dzero{0.00}$^{\circ}$) & 95.47\,{\scriptsize GC} (\dpos{3.42}$^{\dagger}$) & 84.20\,{\scriptsize GC} (\dpos{7.53}$^{\dagger}$) & 98.00\,{\scriptsize D} (\dpos{1.36}$^{\dagger}$) & 98.14\,{\scriptsize NC} (+0.64) \\
DBLP-Scholar & 92.10\,{\scriptsize C} (+0.18) & 93.05\,{\scriptsize C} (+0.48) & 92.27\,{\scriptsize GC} (\dpos{1.81}$^{\dagger}$) & 92.84\,{\scriptsize D} (\dpos{2.63}$^{\dagger}$) & 89.65\,{\scriptsize NC} (+0.67) & 91.26\,{\scriptsize NC} (\dpos{1.14}$^{\dagger}$) & 90.17\,{\scriptsize DC} (\dpos{1.26}$^{\dagger}$) \\
Dn7 Walmart-Amazon & 75.71\,{\scriptsize DP} (\dzero{0.00}$^{\circ}$) & 77.17\,{\scriptsize C} ($-$0.05) & 73.09\,{\scriptsize C} ($-$1.28) & 42.41\,{\scriptsize NS} (\dpos{2.00}$^{\dagger}$) & 68.60\,{\scriptsize NS} (+0.20) & 74.68\,{\scriptsize DP} (\dzero{0.00}$^{\circ}$) & 77.72\,{\scriptsize DP} (\dzero{0.00}$^{\circ}$) \\
billiger.de 100un & 72.54\,{\scriptsize DP} (\dzero{0.00}$^{\circ}$) & 71.74\,{\scriptsize D} (+0.02) & 64.93\,{\scriptsize E} (\dneg{2.98}$^{\dagger}$) & 45.62\,{\scriptsize E} (\dpos{1.82}$^{\dagger}$) & 54.76\,{\scriptsize C} (\dpos{1.50}$^{\dagger}$) & 69.02\,{\scriptsize DP} (\dzero{0.00}$^{\circ}$) & 64.55\,{\scriptsize NS} ($-$1.03) \\
Semi-HETER & 94.64\,{\scriptsize E} (\dpos{4.09}$^{\dagger}$) & 95.82\,{\scriptsize GS} (\dpos{1.70}$^{\dagger}$) & 93.85\,{\scriptsize GS} (+0.43) & 94.77\,{\scriptsize GC} (+0.54) & 89.87\,{\scriptsize D} ($-$1.38) & 94.27\,{\scriptsize D} ($-$0.57) & 94.16\,{\scriptsize D} ($-$0.24) \\
\midrule
Mean F1 & \underline{87.15} & \best{88.41} & 85.74 & 74.06 & 78.09 & 86.03 & 87.31 \\
$\Delta$ Mean $-$ DP & \dpos{0.72}$^{\dagger}$ & +0.28 & $-$0.23 & \dpos{1.34}$^{\dagger}$ & \dpos{0.98}$^{\dagger}$ & +0.10 & +0.15 \\
$\Delta$ Best on test $-$ DP & +1.05$^{\circ}$ & +0.63$^{\circ}$ & +0.67$^{\circ}$ & +1.57$^{\circ}$ & +1.65$^{\circ}$ & +0.54$^{\circ}$ & +1.56$^{\circ}$ \\
Rank of DP & 3 & 1 & 2 & 2 & 3 & 1 & 2 \\
\bottomrule
\end{tabular*}
\end{table*}

\textbf{Cost:} Each decision model incurs a per-pair cost of at most 13\% of that of Sol, the cheapest frontier LLM. PPLX has both the highest mean F1 and the lowest cost of the decision models (Table~\ref{tab:cost}). Projected linearly from our test sets and using the default prompt, one million pairs cost 7.1 USD with PPLX against 753 to 2,715 USD with the frontier LLMs. The cost of a decision model depends on the number of input tokens per pair, which varies with the tokenizer and the request format of each model, and on the price per token, whereas the generative LLMs are also billed for their output tokens. On our benchmarks, PPLX stays 2.7 points below the mean F1 of the frontier LLMs at less than 1\% of their cost.

\textbf{Throughput:} Through the APIs, the decision models process record pairs 1.4 to 1.9 times as fast as Luna Gen.\ (Table~\ref{tab:throughput}), and each frontier LLM is slower than every decision model. For matching large numbers of pairs through an API, a decision model can thus also be faster than querying a generative LLM.

\textbf{Local inference:} We locally run PPLX, its backbone Qwen3.8-27B, and Ditto on one H100 NVL GPU, and the generative mixture-of-experts model GLM-5.3-Flash (320B parameters, 18B active) on four, each on the full test sets in three runs (see lower part of Table~\ref{tab:throughput}).\footnote{\raggedright The exact inference configurations are in our repository.} On one GPU, PPLX reaches the F1 of GLM-5.3-Flash on four GPUs at 86\% of its throughput. Compared with its generatively prompted backbone, PPLX processes 13\% fewer pairs per second with an F1 score 14.8 points higher. Ditto is more than 100 times as fast as these models, but it requires task-specific training data, stays 11 F1 points below PPLX when trained on 2,000 Sol-labeled pairs per benchmark (Section~\ref{sec:labeled}), and generalizes poorly to unseen entities~\cite{peeters2024wdc,peeters2025entity}. On local hardware, PPLX is thus the better choice without training data, whereas a fine-tuned matcher such as Ditto remains preferable for very many pairs under tight time limits, provided that training data exists and few entities are unseen.

\section{Robustness}\label{sec:robustness}
This section measures how the wording of the prompt (Section~\ref{sec:prompts}), the serialization of the records (Section~\ref{sec:serialization}), and perturbations of the records (Section~\ref{sec:perturbations}) affect the decision models. %

\subsection{Prompt Sensitivity}\label{sec:prompts}
\begin{table*}[t]
\centering
\caption{Robustness to Input Variations: F1 difference to DP (pp, 3-run mean $\pm$ SD), significant ($^{\dagger}$, blue/red: gain/loss). $^{*}$Four benchmarks.}\label{tab:inputs}
\fontsize{8}{10}\selectfont
\setlength{\tabcolsep}{3pt}
\renewcommand{\arraystretch}{1.12}
\begin{tabular*}{\textwidth}{@{\extracolsep{\fill}}lrrrrrr|r@{}}
\toprule
Input variant & Jev 1.13 & PPLX v1.2 & Clef & Clef-flash & Kev & \shortstack[r]{GPT-6\\Luna Dec.} & \shortstack[r]{GPT-6\\Luna Gen.} \\
\midrule
DP (F1) & 86.43\,$\pm$0.11 & 88.13\,$\pm$0.03 & 85.97\,$\pm$0.00 & 72.72\,$\pm$0.00 & 77.11\,$\pm$0.04 & 85.93\,$\pm$0.00 & 87.16\,$\pm$0.04 \\
\midrule
\multicolumn{8}{@{}l}{\textit{Serialization}} \\
Values only & \dneg{2.35}\,$\pm$0.13$^{\dagger}$ & \dneg{0.88}\,$\pm$0.13$^{\dagger}$ & \dneg{3.31}\,$\pm$0.00$^{\dagger}$ & \dpos{1.83}\,$\pm$0.00$^{\dagger}$ & $-$0.48\,$\pm$0.07 & \dneg{1.27}\,$\pm$0.00$^{\dagger}$ & \dneg{1.98}\,$\pm$0.03$^{\dagger}$ \\
JSON & +0.07\,$\pm$0.21 & +0.12\,$\pm$0.03 & $-$0.33\,$\pm$0.00 & \dneg{2.31}\,$\pm$0.00$^{\dagger}$ & \dneg{3.10}\,$\pm$0.03$^{\dagger}$ & $-$0.53\,$\pm$0.00 & +0.17\,$\pm$0.08 \\
Attribute pairs & \dneg{2.62}\,$\pm$0.14$^{\dagger}$ & $-$0.25\,$\pm$0.08 & \dneg{1.74}\,$\pm$0.00$^{\dagger}$ & $-$0.05\,$\pm$0.00 & \dneg{3.92}\,$\pm$0.07$^{\dagger}$ & \dneg{5.84}\,$\pm$0.00$^{\dagger}$ & \dneg{2.08}\,$\pm$0.07$^{\dagger}$ \\
Omit missing values & $-$0.05\,$\pm$0.11 & \dpos{0.37}\,$\pm$0.09$^{\dagger}$ & $-$0.22\,$\pm$0.00 & \dneg{0.72}\,$\pm$0.00$^{\dagger}$ & \dpos{0.91}\,$\pm$0.15$^{\dagger}$ & +0.15\,$\pm$0.00 & $-$0.23\,$\pm$0.11 \\
Mean over variants & \dneg{1.24}$^{\dagger}$ & $-$0.16 & \dneg{1.40}$^{\dagger}$ & $-$0.31 & \dneg{1.65}$^{\dagger}$ & \dneg{1.87}$^{\dagger}$ & \dneg{1.03}$^{\dagger}$ \\
\midrule
\multicolumn{8}{@{}l}{\textit{Perturbation}} \\
Entry swap & +0.13\,$\pm$0.15 & \dpos{0.62}\,$\pm$0.06$^{\dagger}$ & $-$0.32\,$\pm$0.00 & $-$0.69\,$\pm$0.00 & \dpos{1.65}\,$\pm$0.05$^{\dagger}$ & +0.09\,$\pm$0.00 & \dneg{0.79}\,$\pm$0.19$^{\dagger}$ \\
Reverse attribute order & +0.24\,$\pm$0.24 & \dpos{0.56}\,$\pm$0.07$^{\dagger}$ & +0.05\,$\pm$0.00 & +0.23\,$\pm$0.00 & +0.06\,$\pm$0.10 & +0.12\,$\pm$0.00 & $-$0.15\,$\pm$0.16 \\
Lowercase values$^{*}$ & $-$0.46\,$\pm$0.46 & +0.09\,$\pm$0.05 & $-$0.27\,$\pm$0.00 & \dpos{0.88}\,$\pm$0.00$^{\dagger}$ & +0.10\,$\pm$0.13 & +0.05\,$\pm$0.01 & +0.26\,$\pm$0.08 \\
Attribute deletion & \dneg{3.38}\,$\pm$0.08$^{\dagger}$ & \dneg{2.23}\,$\pm$0.05$^{\dagger}$ & \dneg{3.20}\,$\pm$0.00$^{\dagger}$ & \dneg{2.04}\,$\pm$0.00$^{\dagger}$ & \dneg{2.20}\,$\pm$0.03$^{\dagger}$ & \dneg{3.68}\,$\pm$0.00$^{\dagger}$ & \dneg{3.13}\,$\pm$0.09$^{\dagger}$ \\
Title typos & $-$0.29\,$\pm$0.06 & $-$0.31\,$\pm$0.06 & \dneg{0.71}\,$\pm$0.00$^{\dagger}$ & \dneg{0.85}\,$\pm$0.00$^{\dagger}$ & \dneg{1.08}\,$\pm$0.06$^{\dagger}$ & $-$0.49\,$\pm$0.00 & \dneg{0.96}\,$\pm$0.07$^{\dagger}$ \\
Mean over variants & \dneg{0.75}$^{\dagger}$ & $-$0.25 & \dneg{0.89}$^{\dagger}$ & \dneg{0.49}$^{\dagger}$ & $-$0.29 & \dneg{0.78}$^{\dagger}$ & \dneg{0.95}$^{\dagger}$ \\
\bottomrule
\end{tabular*}
\end{table*}
Each model receives the default prompt (Figure~\ref{fig:prompt-b0}) and nine other zero-shot prompts that differ only in the instruction text. Four MatchGPT prompts come from Peeters et al.~\cite{peeters2025entity}: \textit{general simple} (GS) and \textit{general complex} (GC) ask whether two entity descriptions match or refer to the same real-world entity, and \textit{domain simple} (DS) and \textit{domain complex} (DC) specifically name products or publications instead of the generic term \textit{entity}. Two prompts (NS, NC) follow Narayan et al.~\cite{narayan2022can} and ask whether Record~A and Record~B are the same or equivalent. Three prompts reword the default prompt: \textit{Domain} (D) names the attributes that distinguish products or publications, \textit{Conservative} (C) asks to be conservative about accepting matches, and \textit{Evidence} (E) weighs agreement and conflict in distinguishing attributes.

We average the test F1 of each prompt over three runs on each of the eight benchmarks. To test whether selecting a prompt per benchmark pays off, we additionally select for each model and benchmark the prompt with the highest F1 on the validation set of the benchmark, measured in one run against the benchmark labels, and report the F1 of this prompt on the test set. The exact wording of all prompts and the corresponding F1 scores for each model are available in the project repository.

Table~\ref{tab:prompt-selection} shows the test F1 of the prompts selected on the validation sets. Tuning the prompt has little impact on the decision models. Used on all eight benchmarks, only general simple and conservative significantly exceed the default prompt, both for Kev, by 0.87 and 0.59 points. As a fixed prompt, the default prompt ranks first to third among the ten prompts in mean F1 for every model. Choosing the prompt separately for each benchmark on its validation set raises the mean F1 significantly for Jev, Kev, and Clef-flash, by 0.7 to 1.3 points. The standard deviation of the mean F1 over the ten prompts stays below 1 point for Jev, PPLX, and Clef and reaches 1.8 to 2.8 points for Luna Dec., Kev, and Clef-flash. For Jev, PPLX, Clef, and Luna Dec., most of this variation stems from billiger.de, where the ten prompts span 10 to 20 points. Most of the tested differences to the default prompt are thus not significant, and other prompts and a selection per benchmark raise the mean F1 by at most 1.3 points, so the default prompt is a good starting point for all six decision models on our benchmarks.

\subsection{Input Serialization}\label{sec:serialization}
In the default prompt, we replace the default record serialization with four alternatives. \textit{Values only} concatenates the attribute values of each record without attribute names, \textit{JSON} encodes both records as one JSON object, \textit{attribute pairs} writes each line as a JSON object with the name of one attribute and its values in both records, and \textit{omit missing values} drops empty attributes from the default serialization.

Table~\ref{tab:inputs} shows that the serialization has little impact on the mean F1 of Jev, PPLX, Clef, and Luna Dec.\ as long as attribute names are kept and the two records are passed in sequence: JSON and omitting missing values change it by at most 0.53 points. Clef-flash and Kev, in contrast, lose 2.3 and 3.1 points with JSON. Dropping the attribute names (values only) or interleaving the records attribute by attribute (attribute pairs) lowers F1 significantly for four of the six decision models each, by up to 5.8 points, although Clef-flash gains 1.8 points with values only. These losses are largest on Walmart-Amazon and Dn7 Walmart-Amazon, where values only and attribute pairs cost Jev, PPLX, Clef, and Luna Dec.\ 4.7 and 5.7 points on average, against 1.1 and 1.6 on the other six benchmarks. For these four models, the default serialization is thus a good choice: no alternative raises their mean F1 by more than 0.4 points.

\subsection{Perturbations}\label{sec:perturbations}

\begin{table*}[t]
\centering
\caption{Performance of Model Cascades at $t=0.05$ with DP: F1 (\%) and cost on the test sets, 3-run mean $\pm$ SD. Bold: best per group. $^{\dagger}$: significant vs.\ the model alone, in $\Delta$ random and $\Delta$ similar vs.\ that baseline. $^{\ddagger}$: significant vs.\ + Sol.}\label{tab:cascade}
\fontsize{8}{10}\selectfont
\setlength{\tabcolsep}{2.5pt}
\renewcommand{\arraystretch}{1.12}
\renewcommand{\score}[2]{\mbox{#1{\fontsize{6.5}{8}\selectfont$\pm$#2}}}
\begin{tabular*}{\textwidth}{@{\extracolsep{\fill}}lrrrrrrrrrrrrr@{}}
\toprule
Configuration & Abt-Buy & Walmart & WDC & DBLP-ACM & DBLP-Sch. & Dn7 & billiger & Semi-H. & Mean & $\Delta$ alone & $\Delta$ random & $\Delta$ similar & Cost \\
\midrule
\multicolumn{14}{@{}l}{\textit{Jev 1.13}: 22.6\,\% of the pairs deferred} \\
Alone & 90.91 & 87.51 & 85.30 & 96.97 & 91.93 & 75.71 & 72.54 & 90.54 & \score{86.43}{0.11} & -- & -- & -- & \$0.73 \\
+ Sol & 96.49$^{\dagger}$ & 93.72$^{\dagger}$ & 91.51$^{\dagger}$ & 97.94$^{\dagger}$ & 93.11$^{\dagger}$ & 80.04$^{\dagger}$ & 77.89$^{\dagger}$ & 95.74$^{\dagger}$ & \score{90.80}{0.13} & \dpos{4.37}$^{\dagger}$ & \dpos{3.03}$^{\dagger}$ & \dpos{1.80}$^{\dagger}$ & \$7.23 \\
+ Ensemble & 96.75$^{\dagger}$ & 93.52$^{\dagger}$ & 91.41$^{\dagger}$ & 98.05$^{\dagger}$ & 93.78$^{\dagger}$ & 81.69$^{\dagger}$ & 79.47$^{\dagger}$ & 95.74$^{\dagger}$ & \best{\score{91.30}{0.23}}$^{\ddagger}$ & \dpos{4.87}$^{\dagger}$ & \dpos{3.44}$^{\dagger}$ & \dpos{2.05}$^{\dagger}$ & \$40.32 \\
\midrule
\multicolumn{14}{@{}l}{\textit{PPLX v1.2}: 12.7\,\% of the pairs deferred} \\
Alone & 93.81 & 90.47 & 87.01 & 98.11 & 92.57 & 77.22 & 71.71 & 94.12 & \score{88.13}{0.03} & -- & -- & -- & \$0.21 \\
+ Sol & 96.49$^{\dagger}$ & 93.96$^{\dagger}$ & 91.58$^{\dagger}$ & 97.94 & 93.15 & 79.65 & 77.98$^{\dagger}$ & 95.74$^{\dagger}$ & \score{90.81}{0.13} & \dpos{2.68}$^{\dagger}$ & \dpos{2.32}$^{\dagger}$ & \dpos{1.75}$^{\dagger}$ & \$4.40 \\
+ Ensemble & 96.82$^{\dagger}$ & 93.52$^{\dagger}$ & 91.48$^{\dagger}$ & 98.05 & 93.79$^{\dagger}$ & 81.37$^{\dagger}$ & 79.49$^{\dagger}$ & 95.74$^{\dagger}$ & \best{\score{91.28}{0.19}}$^{\ddagger}$ & \dpos{3.15}$^{\dagger}$ & \dpos{2.69}$^{\dagger}$ & \dpos{2.10}$^{\dagger}$ & \$25.74 \\
\midrule
\multicolumn{14}{@{}l}{\textit{Clef}: 18.7\,\% of the pairs deferred} \\
Alone & 90.87 & 88.02 & 85.48 & 97.24 & 90.46 & 74.37 & 67.91 & 93.42 & \score{85.97}{0.00} & -- & -- & -- & \$2.91 \\
+ Sol & 96.49$^{\dagger}$ & 93.96$^{\dagger}$ & 91.58$^{\dagger}$ & 97.94 & 93.28$^{\dagger}$ & 79.65$^{\dagger}$ & 77.87$^{\dagger}$ & 95.74$^{\dagger}$ & \score{90.81}{0.13} & \dpos{4.84}$^{\dagger}$ & \dpos{3.84}$^{\dagger}$ & \dpos{1.75}$^{\dagger}$ & \$8.76 \\
+ Ensemble & 96.75$^{\dagger}$ & 93.52$^{\dagger}$ & 91.48$^{\dagger}$ & 98.05 & 93.91$^{\dagger}$ & 81.37$^{\dagger}$ & 79.46$^{\dagger}$ & 95.74$^{\dagger}$ & \best{\score{91.28}{0.22}}$^{\ddagger}$ & \dpos{5.31}$^{\dagger}$ & \dpos{4.23}$^{\dagger}$ & \dpos{2.01}$^{\dagger}$ & \$38.61 \\
\midrule
\multicolumn{14}{@{}l}{\textit{GPT-6 Luna Dec.}: 11.3\,\% of the pairs deferred} \\
Alone & 91.03 & 84.58 & 86.54 & 96.64 & 90.12 & 74.68 & 69.02 & 94.83 & \score{85.93}{0.00} & -- & -- & -- & \$1.24 \\
+ Sol & 96.49$^{\dagger}$ & 93.96$^{\dagger}$ & 91.05$^{\dagger}$ & 97.55$^{\dagger}$ & 93.11$^{\dagger}$ & 79.19$^{\dagger}$ & 77.81$^{\dagger}$ & 96.07 & \score{90.66}{0.12} & \dpos{4.73}$^{\dagger}$ & \dpos{4.08}$^{\dagger}$ & \dpos{3.20}$^{\dagger}$ & \$5.11 \\
+ Ensemble & 96.75$^{\dagger}$ & 93.52$^{\dagger}$ & 90.99$^{\dagger}$ & 97.65$^{\dagger}$ & 93.80$^{\dagger}$ & 80.83$^{\dagger}$ & 79.31$^{\dagger}$ & 96.07 & \best{\score{91.12}{0.21}}$^{\ddagger}$ & \dpos{5.19}$^{\dagger}$ & \dpos{4.47}$^{\dagger}$ & \dpos{3.54}$^{\dagger}$ & \$25.02 \\
\midrule
\multicolumn{14}{@{}l}{\textit{Partners alone, all test pairs}} \\
Sol & 96.49 & 93.72 & 91.58 & 97.94 & 93.11 & 79.90 & 77.87 & 95.74 & \score{90.79}{0.15} & -- & -- & -- & \$22.33 \\
Ensemble & 96.75 & 93.52 & 91.48 & 98.05 & 93.78 & 81.69 & 79.46 & 95.74 & \best{\score{91.31}{0.22}} & -- & -- & -- & \$142.24 \\
\bottomrule
\end{tabular*}
\end{table*}

We perturb the records in five ways that keep the identity of the entities. \textit{Entry swap} exchanges Record~A and Record~B, and \textit{reverse attribute order} reverses the order of the attributes in both records. The \textit{lowercase values} perturbation converts all string values to lowercase and runs only on Dn7 Walmart-Amazon, WDC Products, billiger.de, and Semi-HETER, because the other four benchmarks are lowercase already. \textit{Attribute deletion} removes one randomly chosen non-empty attribute other than the title from each record, and the \textit{title typos} perturbation swaps two different adjacent internal characters in 10\% of the alphabetic title tokens with at least four characters, and in at least one such token per title, if any. All random choices are seeded per pair and record, so the three runs use the same perturbed records.

Table~\ref{tab:inputs} shows that the decision models tolerate changes in how the records are written or ordered, but lose F1 when an attribute is deleted. Entry swap, reverse attribute order, and lowercase values change the mean F1 of each decision model by at most 1.7 points and never lower it significantly, and title typos cost at most 1.1 points. Attribute deletion, in contrast, lowers F1 significantly for all six decision models and Luna Gen., by 2.0 to 3.7 points. 

Across the three studies, Jev, PPLX, Clef, and Luna Dec.\ need little tuning of the prompt and serialization on our benchmarks and cope with small changes in how the records are written or ordered.

\section{Model Cascades}\label{sec:uncertainty}\label{sec:cascade}

Decision models cost substantially less than the frontier LLMs (Section~\ref{sec:cost-throughput}). This section and Section~\ref{sec:labeled} investigate how far F1 can be improved at modest additional cost. This section evaluates model cascades that defer the record pairs on which the decision model is less certain to frontier LLMs.

A model cascade lets a cheap model decide the inputs on which it is confident and defers the remaining inputs to a more expensive model~\cite{chen2023frugalgpt,gupta2024cascades}. Previous work has applied this approach to decision models~\cite{li2026jevcascade,rao2026rubric} and product linking~\cite{wang2026retrieve}. Our cascade uses the match probabilities of the decision models (Section~\ref{sec:calibration}) and defers the less certain pairs to an ensemble of GPT-6.1 Sol, Claude Opus 5.5, and Gemini 4 Argon, which decide by majority vote.

\textbf{Deferral rule:} Following Chow's rule~\cite{chow1970optimum}, the decision model with DP keeps the pairs on which its confidence $\max(p,1-p)$ is high and defers all other pairs. It keeps a pair if $\min(p,1-p) \le t$ for a deferral threshold~$t$ and then predicts a match if $p \ge 0.5$. We fix $t=0.05$ in advance, so the decision model keeps the pairs with $p \le 0.05$ or $p \ge 0.95$. A larger $t$ keeps more pairs. In Section~\ref{sec:threshold} we explore tuning $t$ on labeled pairs.

\textbf{Settings:} Kev and Clef-flash reach the lowest mean F1 (Section~\ref{sec:overall}), and their match probabilities have the highest calibration errors and the lowest resolution (Section~\ref{sec:calibration}), so we run the cascade only for the other four decision models. The cost of a cascade adds the cost of the decision model for all test pairs and the cost of the partner for the deferred pairs, so it grows with the share of deferred pairs. As references, each ensemble member and the majority vote also decide all test pairs alone. We employ two baselines to assess whether match probabilities effectively identify record pairs for deferral: 1) a random baseline that defers the same number of randomly selected pairs per benchmark as the cascade, and 2) a similarity baseline that defers the same number of pairs based on their proximity to a decision boundary defined by surface-form similarity.
For the similarity baseline, we determine a decision boundary for each benchmark using its validation set. We compute two similarity scores for each record pair: the Jaccard similarity between the records' token sets and the cosine similarity between their embeddings generated by text-embedding-3-small. For each similarity measure, we sort the validation pairs in descending order of similarity and set the threshold to the similarity score of the $k$-th pair, where $k$ is the number of matching pairs in the validation set. On the test set, the baseline defers the same number of pairs as the cascade, selecting those closest to the thresholds by alternating between pairs above and below each threshold and between the two similarity measures. All cascades cover three runs and pair run~$r$ of the decision model with run~$r$ of each partner. Table~\ref{tab:cascade} reports the results.

\textbf{Effectiveness:} The cascades defer 11\% to 23\% of the pairs and raise the mean F1 of all four decision models significantly, by 3.2 to 5.3 points with the ensemble and by 2.7 to 4.8 points with Sol as the only partner. Each cascade also exceeds the random baseline significantly, by 2.3 to 4.5 points, so the match probabilities select pairs on which the partner helps more than on random pairs. Each cascade exceeds the similarity baseline significantly as well, by 1.8 to 3.5 points, likely because the pairs near the similarity threshold contain 45\% to 67\% of the errors of each decision model, whereas the pairs that the cascades defer contain 84\% to 99\%. The gain with the ensemble is largest on the product benchmarks, up to 11.5 points on billiger.de, where the four decision models alone reach their lowest F1 (Table~\ref{tab:baseline}). With Sol instead of the ensemble, the cascades keep most of this gain: the ensemble adds 0.46 to 0.50 points, a difference that is significant for all four models, at four to six times the cost.

\textbf{Cost:} With Sol as partner, the cascades show no significant F1 difference to Sol alone, with 95\% intervals between $-0.34$ and $+0.12$ points, at a cost 2.5 to 5.1 times lower. With the ensemble, they stay within 0.2 points of the ensemble alone at a cost 3.5 to 5.7 times lower, a loss that is significant only for Luna Dec. The decision models decide 77\% to 89\% of the pairs themselves. On these pairs, they err on 0.03\% to 0.52\% and Sol on 0.04\% to 0.48\%, a difference of at most 0.04 percentage points per model, which helps explain the small observed differences in mean F1 between the cascades and Sol.
The deferred pairs hold 77\% to 99\% of the errors of Sol, so further F1 gains would have to come mostly from these pairs, for example through manual review. %

\section{Matching using Development Data}\label{sec:labeled}
\newcommand{\HandwrittenLossPPLX}{2.1}
\newcommand{\HandwrittenLossJev}{1.3}
\newcommand{\HandwrittenLossPPLXbilliger}{9.6}
\newcommand{\HandwrittenRuns}{1}%
This section investigates to what extent matching performance can be improved using a development set of record pairs labeled as matches or non-matches. Traditionally, manually labeling such data has been a major cost driver in entity matching projects. In \cite{steiner2026labeling}, we demonstrated that using frontier LLMs to label development sets can substantially reduce labeling costs. Following this approach, we use GPT-6.1 Sol to label the record pairs in the development set. We investigate three ways of leveraging these labeled pairs: (1) tuning decision and deferral thresholds, (2) providing demonstrations for in-context learning, and (3) using an LLM to derive data set-specific matching rules.

\textbf{Development sets:}  We collect all unique records from each benchmark and build a pool of candidate pairs for the benchmark using embedding-based blocking. From this pool, we select pairs for labeling by similarity search and active learning as described in our earlier study~\cite{steiner2026labeling}, while excluding every pair of the benchmark validation and test sets. We use GPT-6.1 Sol together with the DP prompt to label a development set containing 200 pairs and a development set containing 2,000 pairs for each benchmark. To validate the quality of the LLM-labeled sets, we compare the labels of the pairs in the 2,000 sets that overlap with the original benchmark training sets and find that the Sol-generated labels reach an F1 of 81 to 95 depending on the benchmark. The F1 is 94 to 95 on Abt-Buy, Walmart-Amazon, and DBLP-ACM, 90.5 on WDC Products, where Sol labels fewer pairs as matches than the benchmark (recall 84\%), and 88.9 on Dn7 Walmart-Amazon. On the three benchmarks with the lowest F1, the disagreements differ in direction: on Semi-HETER (F1 81.4), Sol labels more pairs as matches than the benchmark (precision 72\%, recall 94\%), on billiger.de (82.0), it labels fewer (precision 98\%, recall 70\%), and on DBLP-Scholar (84.7), it deviates in both directions (precision 89\%, recall 81\%).
On WDC Products and billiger.de, whose test sets contain unseen entities, the development sets share no record with the test sets. Labeling with Sol costs an estimated 0.15 USD per benchmark for the 200 sets and 1.55 USD per benchmark for the 2,000 sets, in total 1.2 and 12.4 USD for the eight benchmarks. 

\subsection{Threshold Tuning}\label{sec:threshold}
\textbf{Decision threshold:} All results so far decided at the default threshold of 0.5. The simplest use of labeled development pairs is to tune this decision threshold, which shifts the decision boundary but leaves the probabilities unchanged. For each decision model with DP, we choose the threshold that maximizes F1 on the Sol labels of the 2,000 sets, once for each benchmark and once for all benchmarks together. We apply both thresholds to the three test runs and report the mean F1 over the benchmarks in Table~\ref{tab:thresholds}. The values per benchmark are in our repository.

\begin{table}[t]
\centering\small
\setlength{\tabcolsep}{2.5pt}
\caption{Threshold Tuning Results: Test F1 (\%) at $t=0.5$ and $\Delta$F1 ($\pm$ SD) for $t$ tuned on the 2,000 sets. $^{\dagger}$: significant.}\label{tab:thresholds}
\renewcommand{\score}[2]{\mbox{#1{\fontsize{6.5}{8}\selectfont$\pm$#2}}}
\begin{tabular}{@{}lrrrr@{}}
\toprule
 & F1 at & $t$ per & \multicolumn{2}{c}{$\Delta$F1 with $t$ tuned} \\
\cmidrule(l){4-5}
Model & $t=0.5$ & model & per bench. & per model \\
\midrule
Jev 1.13 & 86.43 & 0.49 & \score{\dpos{0.46}}{0.11}$^{\dagger}$ & \score{\dpos{0.16}}{0.12} \\
PPLX v1.2 & 88.13 & 0.54 & \score{\dpos{0.02}}{0.10} & \score{\dneg{0.05}}{0.05} \\
Clef & 85.97 & 0.64 & \score{\dneg{0.84}}{0.00}$^{\dagger}$ & \score{\dneg{0.67}}{0.00}$^{\dagger}$ \\
Clef-flash & 72.72 & 0.78 & \score{\dpos{4.56}}{0.00}$^{\dagger}$ & \score{\dpos{5.41}}{0.00}$^{\dagger}$ \\
Kev & 77.11 & 0.75 & \score{\dpos{1.58}}{0.01}$^{\dagger}$ & \score{\dpos{0.22}}{0.12} \\
GPT-6 Luna Dec. & 85.93 & 0.82 & \score{\dpos{1.07}}{0.01}$^{\dagger}$ & \score{\dpos{1.62}}{0.01}$^{\dagger}$ \\
\bottomrule
\end{tabular}
\end{table}

Table~\ref{tab:thresholds} shows the results. A tuned threshold helps the three models that overestimate the match probability most. Clef-flash and Luna Dec.\ gain with both kinds of threshold, by up to 5.4 and 1.6 points, and Kev gains 1.6 points with thresholds per benchmark. Their shared thresholds of 0.75 to 0.82 lie well above the default, as their reliability diagrams suggest (Figure~\ref{fig:reliability}). For PPLX, Clef, and Jev, which have the lowest calibration errors, tuning changes F1 by less than one point, and Clef even loses up to 0.84 points. The shared threshold of Jev is 0.49, close to the default, although Deu{\ss}er et al.~\cite{deusser2026evaluating} recommend tuning it, and thresholds per benchmark raise its F1 significantly, by 0.46 points. This is in line with theory: for calibrated probabilities, the threshold that maximizes F1 equals half of the maximal F1~\cite{lipton2014optimal}, which is 0.44 for an F1 of 88\% and thus close to the default. A tuned threshold thus compensates for a high calibration error but leaves the resolution unchanged (Section~\ref{sec:calibration}). It helps Clef-flash, Kev, and Luna Dec.\ and adds little for PPLX, Clef, and Jev, whose calibration errors are lowest, so users of these three models can likely keep the default.

\textbf{Number of labels:} With 200 instead of 2,000 labeled pairs per benchmark, the shared threshold still reaches 5.0 of the 5.4 points for Clef-flash and 1.0 of the 1.6 points for Luna Dec. Thresholds per benchmark, in contrast, likely overfit the smaller sets: they no longer raise F1 significantly for any model except Clef-flash and cost Clef 1.9 points. A few hundred labeled pairs thus likely suffice for one shared threshold per model, but not for thresholds per benchmark.

\textbf{Deferral threshold:} The development set can also set the deferral threshold $t$ of the cascades of Section~\ref{sec:cascade}. We choose $t$ as in selective classification~\cite{geifman2017selective}. The selective risk is the share of wrong decisions among the kept pairs. For a target risk~$\epsilon$, $t$ is the largest threshold whose selective risk on the pooled Sol-labeled 2,000 sets, measured against the Sol labels, is at most~$\epsilon$. We use the target risks 0.25\%, 0.5\%, and 1\%. Labeling the 2,000 set of a benchmark with Sol costs an estimated 1.55 USD, based on the mean cost per call. This estimate covers only labeling the selected pairs, as the selection reuses the labeling runs of our earlier study~\cite{steiner2026labeling}. At a target risk of 0.5\%, these labels bring no F1 gain, but they save cost for three of the four models (Table~\ref{tab:labeled} for PPLX). The tuned $t$ of Jev, PPLX, and Clef defers fewer pairs than $t=0.05$, which lowers the cost of their cascades by a factor of 1.4 to 2.6, at a significant F1 loss of 0.26 to 0.41 points with the ensemble and without a significant F1 change with Sol. In cost, the labels thus pay for themselves after about 1,900 to 3,300 pairs per benchmark with the ensemble and 11,000 to 20,000 with Sol. For Luna Dec., the tuned $t$ of 0.04 is lower than 0.05, so its cascades cost 12\% to 15\% more at about the same F1. The target risk limits the deviation from the Sol labels on the 2,000 sets, not the error on a test set: at 0.5\%, the selective risk on the pooled test sets is 0.36\% to 0.85\% against the benchmark labels, and up to 2.8\% on single benchmarks. The other two target risks trade cost against F1: 0.25\% raises the cost of the cascades by 17\% to 47\% for at most 0.21 points of mean F1, and 1\% lowers the cost by 18\% to 38\% at a loss of up to 0.70 points. On our benchmarks, a target risk of 0.5\% thus avoids the F1 losses of 1\% at a lower cost than 0.25\%.

\subsection{In-Context Learning}\label{sec:labeled-setup}
We investigate three in-context learning approaches: 1) adding demonstrations to the prompt, 2) deriving textual matching rules from the development set using an LLM, and 3) the combination of both. The demonstrations and the rules are appended to the instructions section of the DP prompt (Figure~\ref{fig:prompt-b0}). We evaluate in-context learning for the decision models PPLX and Jev. We split each development set into 80\% of its pairs, from which we derive the rules and select demonstrations, and 20\% on which the LLM checks the rules, with the same share of matches in both splits. The LLM may revise its rules. However, no rule set is rejected. 

\textbf{Generated rules:} We use the generative LLM GPT-6 Astra via OpenAI Codex to derive matching rules. We set no limit on the number or length of the rules. The agent writes 8 to 13 rules per set. As the rules differ between runs, we execute the agent three times per set and report the mean. Our GitHub repository offers all the generated rules for inspection.

\textbf{Demonstrations:} We add the 10 labeled pairs that are most similar to the record pair to be matched as demonstrations to the prompt. We rank the labeled pairs by the BM25 score~\cite{robertson2009probabilistic} of the serialized pairs and by the mean cosine similarity of the embeddings of their records, fuse the two rankings by reciprocal rank fusion~\cite{cormack2009reciprocal}, and take at least two demonstrations from each class. Demonstrations that consist of the same two records as the test pair are removed.

As PLM baseline, Table~\ref{tab:labeled} lists Ditto~\cite{li2020deep} with RoBERTa-base, trained using the hyperparameter settings of our earlier study~\cite{steiner2026labeling} on our 200 and 2,000 sets, and as reported there for the full training set of each benchmark, both as the mean of three seeds. It also lists the cascades of PPLX with DP at $t=0.05$ and with the tuned $t$.

\begin{table*}[t]
\centering
\caption{Test F1 (\%) of methods using the LLM-labeled development sets, 3-run or 3-seed mean ($\pm$ SD). $\Delta$, $^{\dagger}$: vs.\ DP of the same model (Ditto, cascades: vs.\ PPLX with DP). Bold/underl.: best/second without Sol, cascades, training sets. \$/1M: inference only.}\label{tab:labeled}
\fontsize{8}{10}\selectfont
\setlength{\tabcolsep}{2.2pt}
\renewcommand{\arraystretch}{1.12}
\renewcommand{\score}[2]{\mbox{#1{\fontsize{6.5}{8}\selectfont$\pm$#2}}}
\begin{tabular*}{\textwidth}{@{\extracolsep{\fill}}llrrrrrrrrrrr@{}}
\toprule
Method & Labels & Abt-Buy & Walm.-Am. & WDC & DBLP-ACM & DBLP-Sch. & Dn7 & billiger & Semi-H. & Mean & $\Delta$ & \$/1M \\
\midrule
\multicolumn{13}{@{}l}{\textit{No labeled pairs}} \\
PPLX v1.2 (DP) & 0 & 93.81 & \best{90.47} & 87.01 & 98.11 & 92.57 & 77.22 & 71.71 & 94.12 & \score{88.13}{0.03} & -- & \$7.1 \\
Jev 1.13 (DP) & 0 & 90.91 & 87.51 & 85.30 & 96.97 & 91.93 & 75.71 & 72.54 & 90.54 & \score{86.43}{0.11} & -- & \$24.1 \\
GPT-6.1 Sol & 0 & 96.49$^{\dagger}$ & 93.72$^{\dagger}$ & 91.58$^{\dagger}$ & 97.94 & 93.11 & 79.90 & 77.87$^{\dagger}$ & 95.74$^{\dagger}$ & \score{90.79}{0.15} & \dpos{2.66}$^{\dagger}$ & \$753 \\
\midrule
\multicolumn{13}{@{}l}{\textit{PLM Matcher trained on Sol-labeled pairs}} \\
Ditto & 200 & 18.94$^{\dagger}$ & 31.86$^{\dagger}$ & 20.38$^{\dagger}$ & 82.40$^{\dagger}$ & 82.05$^{\dagger}$ & 28.81$^{\dagger}$ & 19.11$^{\dagger}$ & 84.57$^{\dagger}$ & \score{46.01}{2.61} & \dneg{42.12}$^{\dagger}$ & -- \\
Ditto & 2,000 & 89.56$^{\dagger}$ & 83.75$^{\dagger}$ & 56.37$^{\dagger}$ & 97.90 & 90.84$^{\dagger}$ & 60.11$^{\dagger}$ & 46.43$^{\dagger}$ & 93.25 & \score{77.28}{1.88} & \dneg{10.85}$^{\dagger}$ & -- \\
\midrule
\multicolumn{13}{@{}l}{\textit{Decision models using Sol-labeled pairs}} \\
PPLX v1.2 + rules & 200 & 94.91 & 89.07 & 87.15 & 98.23 & \best{93.69}$^{\dagger}$ & 78.46 & 75.78$^{\dagger}$ & 94.84 & \score{89.02}{0.09} & \dpos{0.89}$^{\dagger}$ & \$23.7 \\
PPLX v1.2 + rules & 2,000 & 94.06 & 89.26 & 87.66 & 97.91 & \underline{93.05} & 77.95 & 76.11$^{\dagger}$ & \best{95.85}$^{\dagger}$ & \score{88.98}{0.08} & \dpos{0.85}$^{\dagger}$ & \$26.9 \\
PPLX v1.2 + 10 dem. & 200 & 95.67$^{\dagger}$ & 86.25$^{\dagger}$ & \underline{87.83} & 97.79 & 91.45$^{\dagger}$ & 78.92 & \underline{76.78}$^{\dagger}$ & \underline{95.74}$^{\dagger}$ & \score{88.80}{0.03} & \dpos{0.67}$^{\dagger}$ & \$62.7 \\
PPLX v1.2 + 10 dem. & 2,000 & \underline{97.49}$^{\dagger}$ & \underline{90.35} & 87.62 & \best{98.37} & 92.49 & \best{80.07}$^{\dagger}$ & 76.53$^{\dagger}$ & 94.84 & \underline{\score{89.72}{0.09}} & \dpos{1.59}$^{\dagger}$ & \$67.5 \\
PPLX v1.2 + rules + 10 dem. & 200 & 95.43 & 85.38$^{\dagger}$ & \best{88.24}$^{\dagger}$ & 98.19 & 92.61 & 78.10 & 76.60$^{\dagger}$ & 94.90 & \score{88.68}{0.05} & \dpos{0.55} & \$79.3 \\
PPLX v1.2 + rules + 10 dem. & 2,000 & \best{97.64}$^{\dagger}$ & 90.15 & 87.73 & \underline{98.26} & 92.54 & \underline{79.10} & \best{78.92}$^{\dagger}$ & \best{95.85}$^{\dagger}$ & \best{\score{90.02}{0.11}} & \dpos{1.89}$^{\dagger}$ & \$86.3 \\
Jev 1.13 + rules & 200 & 90.45 & 88.22 & 83.62$^{\dagger}$ & 97.87 & 92.55 & 76.01 & 74.16 & 94.00$^{\dagger}$ & \score{87.11}{0.31} & \dpos{0.68}$^{\dagger}$ & \$59.2 \\
Jev 1.13 + rules & 2,000 & 90.52 & 87.22 & 83.23$^{\dagger}$ & 97.88$^{\dagger}$ & 91.75 & 76.35 & 74.49 & 95.72$^{\dagger}$ & \score{87.14}{0.31} & \dpos{0.71}$^{\dagger}$ & \$66.1 \\
Jev 1.13 + 10 dem. & 200 & 92.62$^{\dagger}$ & 87.75 & 86.11 & 97.61 & 91.65 & 78.08$^{\dagger}$ & 72.40 & 94.23$^{\dagger}$ & \score{87.56}{0.23} & \dpos{1.13}$^{\dagger}$ & \$144 \\
Jev 1.13 + 10 dem. & 2,000 & 93.48$^{\dagger}$ & 89.09 & 85.36 & 97.79$^{\dagger}$ & 92.75$^{\dagger}$ & 77.59 & 72.19 & 94.25$^{\dagger}$ & \score{87.81}{0.08} & \dpos{1.38}$^{\dagger}$ & \$155 \\
Jev 1.13 + rules + 10 dem. & 200 & 92.21 & 88.43 & 83.35$^{\dagger}$ & 98.22$^{\dagger}$ & 92.76$^{\dagger}$ & 77.27 & 74.10 & 94.00$^{\dagger}$ & \score{87.54}{0.14} & \dpos{1.11}$^{\dagger}$ & \$179 \\
Jev 1.13 + rules + 10 dem. & 2,000 & 90.75 & 88.38 & 82.46$^{\dagger}$ & 97.75 & 91.09 & 76.36 & 74.57 & 95.60$^{\dagger}$ & \score{87.12}{0.06} & \dpos{0.69} & \$195 \\
\midrule
\multicolumn{13}{@{}l}{\textit{Model cascades of PPLX v1.2, $t=0.05$ or $t$ tuned on the Sol-labeled 2,000 sets}} \\
+ Sol, $t=0.05$ & 0 & 96.49$^{\dagger}$ & 93.96$^{\dagger}$ & 91.58$^{\dagger}$ & 97.94 & 93.15 & 79.65 & 77.98$^{\dagger}$ & 95.74$^{\dagger}$ & \score{90.81}{0.13} & \dpos{2.68}$^{\dagger}$ & \$146 \\
+ Sol, tuned $t=0.22$ & 2,000 & 96.31$^{\dagger}$ & 93.46$^{\dagger}$ & 90.82$^{\dagger}$ & 98.05 & 93.18 & 80.54$^{\dagger}$ & 77.67$^{\dagger}$ & 95.40 & \score{90.68}{0.07} & \dpos{2.55}$^{\dagger}$ & \$68.0 \\
+ ensemble, $t=0.05$ & 0 & 96.82$^{\dagger}$ & 93.52$^{\dagger}$ & 91.48$^{\dagger}$ & 98.05 & 93.79$^{\dagger}$ & 81.37$^{\dagger}$ & 79.49$^{\dagger}$ & 95.74$^{\dagger}$ & \score{91.28}{0.19} & \dpos{3.15}$^{\dagger}$ & \$852 \\
+ ensemble, tuned $t=0.22$ & 2,000 & 96.72$^{\dagger}$ & 93.18$^{\dagger}$ & 90.73$^{\dagger}$ & 98.16 & 93.77$^{\dagger}$ & 81.56$^{\dagger}$ & 78.66$^{\dagger}$ & 95.40 & \score{91.02}{0.16} & \dpos{2.89}$^{\dagger}$ & \$390 \\
\midrule
\multicolumn{13}{@{}l}{\textit{Benchmark training set}} \\
\textit{Labeled pairs} & & 5,758 & 6,146 & 19,835 & 7,441 & 17,260 & 26,050 & 26,571 & 1,240 & & & \\
Ditto & all & 88.34$^{\circ}$ & 85.66$^{\circ}$ & 71.23$^{\circ}$ & 98.43$^{\circ}$ & 95.64$^{\circ}$ & 72.20$^{\circ}$ & 51.84$^{\circ}$ & 81.61$^{\circ}$ & 80.62 & \dneg{7.51}$^{\circ}$ & -- \\
\bottomrule
\end{tabular*}
\end{table*}

\textbf{Effectiveness:} In-context learning increases the mean F1 of both decision models, significantly in most configurations, by up to 1.9 points for PPLX and 1.4 for Jev (Table~\ref{tab:labeled}). Rules derived from 200 pairs raise it by 0.9 and 0.7 points. Ten retrieved demonstrations help somewhat less than these rules for PPLX with 200 pairs and more in the other three combinations of model and set size, and combining rules and demonstrations adds 0.3 points over the better of the two for PPLX with 2,000 pairs and nothing in the other three, so their gains do not add up. For PPLX, both help most on billiger.de, where DP is weakest: derived rules add 4.1 to 4.4 points there. Rules and demonstrations derived from labeled pairs thus raise the mean F1 of PPLX and Jev by up to about two points. Rules derived from 2,000 instead of 200 pairs add about as much, whereas ten demonstrations retrieved from 2,000 pairs add 0.9 points more for PPLX and 0.3 for Jev. A few hundred labeled pairs thus likely suffice for rules, whereas demonstrations profit from more pairs.

\textbf{Comparison to Ditto:} Even with thousands of labels, Ditto stays below PPLX with DP, and it loses most on unseen products. Trained on our 200 and 2,000 sets, it reaches 46.0 and 77.3, and even on the full benchmark training sets, with up to 26,571 pairs, it reaches 80.6 (Table~\ref{tab:labeled}). On WDC Products and billiger.de, whose test sets contain only unseen products (Section~\ref{sec:setup-datasets}), it stays 16 and 20 points behind PPLX with DP even with the full training sets because PLM-based matchers have a tendency to generalize poorly to unseen entities~\cite{peeters2024wdc,peeters2025entity}. Even without using the development set, PPLX and Jev reach a higher mean F1 than Ditto (see the \textit{No labeled pairs} section of Table~\ref{tab:labeled}).

\textbf{Effect on Cascades:} With DP and the fixed deferral threshold $t=0.05$, which is not tuned on labeled pairs, the cascades of PPLX with Sol and with the ensemble reach 90.81 and 91.28, above every configuration of PPLX or Jev alone with rules or demonstrations. Rules and demonstrations from the 2,000 sets raise the mean F1 of PPLX alone significantly, by 1.9 points, but add almost nothing to its cascades: 0.04 points with the ensemble and 0.08 with Sol at $t=0.05$, both not significant. Rules and demonstrations thus add less F1 to PPLX and Jev than deferring their less certain pairs to a frontier LLM.

\textbf{Cost:} Labeling the pairs costs little, but rules and demonstrations raise the cost of every decision. Labeling the 200 sets of the eight benchmarks with GPT-6.1 Sol costs an estimated 1.2 USD, and deriving the rules with GPT-6 Astra about 18 USD per run for the 200 sets and 44 USD for the 2,000 sets. The longer requests raise the cost of PPLX from 7.1 USD per million pairs with DP (Table~\ref{tab:cost}) to 24 to 27 USD with rules, 63 to 67 USD with ten demonstrations, and 79 to 86 USD with both, at most 12\% of the 753 USD of GPT-6.1 Sol. The cascade of PPLX with Sol (Section~\ref{sec:cascade}) costs about 146 USD per million pairs with $t=0.05$ and 68 USD with the tuned $t$, about as much as PPLX with ten demonstrations. Rules and demonstrations also shift the match probabilities of PPLX toward 0 and 1, so that its cascade with the ensemble defers 7.3\% instead of 12.7\% of the pairs and costs 25\% less. With Sol as partner, the cascade with DP remains cheaper, as rules and demonstrations raise the cost of PPLX for every pair by more than the fewer deferred pairs save. The one-time costs pay off only for larger tasks: counting the labels that each way needs, rules become cheaper than the cascade with $t=0.05$ from about 20,000 pairs per benchmark with the 200 sets and from about 60,000 with the 2,000 sets, at 1.8 F1 points less. For PPLX, rules derived from 200 pairs thus add the least cost per pair among the ways to gain F1 from labeled pairs, whereas a cascade with Sol gains more F1 at a higher cost per pair.

\section{Related Work}\label{sec:relatedwork}

\textbf{Entity Matching:} Entity matching~\cite{Christen2012DataMC,barlaug2021neural} has been studied for over 50 years~\cite{fellegi1969theory}. Over the past decades, research has been dominated by supervised machine learning approaches~\cite{Maggelan2016}, including matchers based on pretrained language models (PLMs)~\cite{li2020deep,peeters2021dual,peeters2022supervised,tu2023unicorn}. Since 2022, research attention has shifted toward employing large language models (LLMs) for entity matching~\cite{narayan2022can,peeters2023using,fan2024costeffective,wang2025match,smith2026opensanctions,zhang2024jellyfish,steiner2025finetuning}, driven by their reduced need for task-specific training data and their robustness to unseen entities~\cite{peeters2025entity}. To reduce the high inference costs and latency of prompted LLMs, researchers have proposed using smaller fine-tuned language models~\cite{zhang2024anymatch,zhang2025deep} and PLM-based matchers trained on LLM-generated labels through knowledge distillation~\cite{steiner2026labeling}. Other work lowers the inference cost of LLM-based matching through learned operator selection~\cite{guo2026carlem}, batching~\cite{ma2026cerag}, early stopping~\cite{chen2026gleam}, or clustering~\cite{fu2025clustering}, selects demonstrations~\cite{pulsone2026beacon,akashi2026diversity} or induces rules~\cite{zhang2026metarule} when labels are scarce, and calibrates fine-tuned PLM matchers~\cite{kamsteeg2025calibration}. %

\textbf{Decision Models:} In September 2026, TypeSafe introduced the interface described in Section~\ref{sec:setup-models} with Jev, the first of its ``System One'' models. Within a month, Clef and Clef-flash, Kev, and Perplexity's Decider followed with open weights, and OpenAI's Decisions API entered public beta. Tang and Zheng~\cite{tang2026evidenceaudit} audit 28 early papers and conclude that the typed outputs of decision models have not yet shown an independent accuracy advantage over the label probabilities of comparable language models, whereas Jev's clearest gains are in latency and cost. Ling et~al.~\cite{ling2026jevwild} analyze 2,170 public GitHub projects that use Jev and find it mostly used to judge attributes of an input and to score or rank, often combining several question types. Decision models have also been applied to schema matching~\cite{li2026jevnexus} and semantic joins~\cite{wang2026jevdb}. Kim~\cite{kim2026plans} formalizes cost-accuracy optimization of semantic queries given calibrated decision-model confidences. %
\begin{table}[t]
\centering
\caption{Best published F1 of trained matchers and PPLX, alone and in the cascade ($t=0.05$). CE: cross-encoder. $^{\ddagger}$Benchmark-tuned.}\label{tab:literature}
\fontsize{7}{8.6}\selectfont
\setlength{\tabcolsep}{2pt}
\renewcommand{\arraystretch}{1.1}
\begin{tabular*}{\columnwidth}{@{\extracolsep{\fill}}llrrrrr@{}}
\toprule
 & \multicolumn{2}{c}{Best trained matcher} & \multicolumn{2}{c}{PPLX, DP} & \multicolumn{2}{c}{PPLX, rules+dem.} \\
\cmidrule(lr){2-3}\cmidrule(lr){4-5}\cmidrule(l){6-7}
Benchmark & Method & F1 & Alone & Casc. & Alone & Casc. \\
\midrule
Abt-Buy & Qwen3-8B CE~\cite{zhang2026beyond} & 94.9 & 93.81 & 96.82 & \best{97.64} & \underline{97.07} \\
Walmart & Qwen3-4B CE~\cite{zhang2026beyond} & \underline{91.6} & 90.47 & \best{93.52} & 90.15 & \best{93.52} \\
WDC & Qwen3-8B gen.~\cite{zhang2026beyond} & 83.9 & 87.01 & \best{91.48} & 87.73 & \underline{91.21} \\
DBLP-ACM & EMTransformer-B~\cite{papadakis2024critical} & \best{99.21} & 98.11 & 98.05 & 98.26 & \underline{98.27} \\
DBLP-Sch. & Jellyfish-13B$^{\ddagger}$~\cite{zhang2024jellyfish} & \best{98.51} & 92.57 & \underline{93.79} & 92.54 & 93.63 \\
Dn7 & Ditto (LLM labels)~\cite{steiner2026labeling} & 76.22 & 77.22 & \underline{81.37} & 79.10 & \best{81.69} \\
billiger & XLM-R~\cite{steiner2026billiger} & 51.36 & 71.71 & \best{79.49} & 78.92 & \underline{79.46} \\
Semi-H. & Ditto (LLM labels)~\cite{steiner2026labeling} & 94.76 & 94.12 & \underline{95.74} & \best{95.85} & \underline{95.74} \\
\midrule
Mean & & 86.31 & 88.13 & \underline{91.28} & 90.02 & \best{91.32} \\
\multicolumn{3}{@{}l}{Above best trained (of 8)} & 3 & 6 & 5 & 6 \\
\bottomrule
\end{tabular*}
\end{table}

\textbf{Evaluating Decision Models:} Using the JEVal benchmark, Duan et~al.~\cite{duan2026generaldecision} find the strongest decision models as accurate as reasoning LLMs and better calibrated than the verbalized probabilities of all generative LLMs, but weaker on decisions requiring specialist knowledge or faithful uncertainty estimation. In System1-Arena, Chen et~al.~\cite{chen2026system1arena} report that the largest decision models match or exceed fast closed LLMs in non-agentic accuracy, with better-calibrated probabilities than these LLMs verbalize. Deu{\ss}er et~al.~\cite{deusser2026evaluating} find Jev's binary probabilities well ranked but poorly placed relative to a fixed 0.5 threshold and suggest threshold tuning. Other preprints report that decisions can follow option names rather than definitions~\cite{sun2026typesafeerrorfree,azizi2026labels}, can depend on the option set and, in choice questions, on the option order~\cite{rajput2026irrelevant,huang2026orderinvariant}, and change with input serialization~\cite{zhang2026tabular}. In JevAdvBench, black-box attacks on the input or answer options flip 46\% of Jev's correct decisions, but attacks that paraphrase, substitute synonyms, or reorder options flip at most 9\%~\cite{hu2026jevadvbench}. On tabular data, paraphrasing affects decision models and LLMs alike~\cite{zhang2026tabular}. 

\textbf{Entity Matching with Decision Models:} System1-Arena~\cite{chen2026system1arena} includes 500 record pairs from the Abt-Buy, DBLP-Scholar, and WDC Products benchmarks but reports results only as part of a five-task domain average. DecisionPivot~\cite{zhuang2026decisionpivot} includes about 1,000 sampled pairs each from Abt-Buy and Amazon-Google among its 35 tasks and reports their AUROC for six decision models, among them Jev, Perplexity Decider, Kev, and Clef. On 117 synthetic cases, 30 of them on entity matching, Singh and Naik~\cite{singh2026compatible} report that Jev, Clef, and Clef-flash rank the cases well but differ in score scale, so that thresholds set on Jev do not transfer. On 12 DeepMatcher test sets, the jlink repository reports F1 values for Jev that exceed the published F1 of fine-tuned Ditto~\cite{li2020deep} only for Abt-Buy and structured Walmart-Amazon.\footnote{\raggedright\footurl{github.com/keltokhy/jlink}} The KevDB repository compares Jev with open Kev models on the full test sets of four DeepMatcher benchmarks and fine-tunes Kev on labeled pairs of the other three.\footnote{\raggedright\footurl{github.com/ElyasEbrahim/KevDB}} %

\textbf{Comparison to Published Results:}\label{sec:literature} Table~\ref{tab:literature} compares our PPLX results with the highest published F1-scores we identified for matchers trained on the respective benchmarks and evaluated on the same test splits. PPLX appears in two configurations, with DP and with rules and ten demonstrations from the 2,000 set, the configuration of Section~\ref{sec:labeled} with the highest mean F1 on the benchmark validation sets, each alone and in the cascade with the ensemble at $t=0.05$. %

The cascade of PPLX with DP exceeds the best trained matcher on six of the eight benchmarks, the five product benchmarks and Semi-HETER (Table~\ref{tab:literature}), as the ensemble alone does (Table~\ref{tab:cascade}), with no significant F1 difference to the ensemble alone and at less than a fifth of its cost. It stays behind the trained matchers on the two bibliographic benchmarks, where the best of them, Jellyfish-13B, was instruction-tuned on training data of DBLP-Scholar~\cite{zhang2024jellyfish}. PPLX alone exceeds the best trained matcher on three benchmarks with DP and on five with rules and demonstrations. Without using the development set, a cascade combining a decision model with an ensemble of frontier LLMs can surpass the highest published F1-scores of benchmark-trained matchers on product datasets, while benchmark-trained matchers stay ahead on bibliographic data.

\section{Conclusion}\label{sec:conclusion}

This paper presented an extensive experimental evaluation of decision models for entity matching. Our experiments cover six recent decision models that we evaluate using eight entity matching benchmarks focusing on different matching challenges. The experiments demonstrate the suitability of decision models for entity matching: without task-specific training data, the strongest decision models achieved F1 scores within a few points of those of frontier LLMs at a fraction of their cost. The decision models PPLX, Clef, and Jev exhibited low calibration errors and proved quite robust to variations in prompt wording and the order of the input. The match probabilities produced by the models enable effective model cascades: by deferring less certain record pairs to frontier LLMs, these cascades achieve F1 scores that do not significantly differ from the scores of the LLMs alone, while being substantially cheaper. In contrast, using a development set to select in-context demonstrations or derive matching rules yielded smaller F1 improvements than employing model cascades.

\section*{Acknowledgment}

The authors used Claude Code with Claude Opus 5.5 and OpenAI Codex with GPT-6 Astra to develop the code of the experiments and to revise the text of the paper. We used OpenAI's image-generation tool to draw Figure~\ref{fig:teaser}. As part of the evaluated methods, GPT-6.1 Sol labels the development pairs of Section~\ref{sec:labeled}, and GPT-6 Astra, accessed through OpenAI Codex, derives the matching rules of that section. The authors reviewed all AI-assisted content and take full responsibility for it. The design of the experiments and the interpretation of the results are entirely the authors' own work.
\balance

\bibliographystyle{IEEEtranS}
\bibliography{bibliography}

\end{document}